%% file: main.tex
\documentclass[runningheads]{llncs}

\usepackage{eccv}

\usepackage{eccvabbrv}

\usepackage{graphicx}
\usepackage{booktabs}

\usepackage[accsupp]{axessibility}  %

\usepackage{hyperref}

\usepackage{orcidlink}
\usepackage{multirow}
\usepackage{multicol}
\usepackage{makecell}
\usepackage{colortbl}
\usepackage{xcolor}
\usepackage{bm}
\usepackage{amsmath}
\usepackage{float}
\usepackage{stfloats}
\usepackage{wrapfig}
\newlength\savewidth\newcommand\shline{\noalign{\global\savewidth\arrayrulewidth
  \global\arrayrulewidth 1pt}\hline\noalign{\global\arrayrulewidth\savewidth}}

\definecolor{mygreen}{RGB}{0, 205, 108}

\DeclareMathOperator*{\argmin}{arg\,min}

\begin{document}

\title{MLLMs-Augmented Visual-Language Representation Learning}

\author{Yanqing Liu\inst{1,2}\textsuperscript{*}  \and
Kai Wang\inst{1}\textsuperscript{*\dag} \and
Wenqi Shao\inst{2} \and
Ping Luo\inst{2,3} \and
Yu Qiao\inst{2} \and
\\
Mike Zheng Shou\inst{1} \and
Kaipeng Zhang\inst{2}\textsuperscript{\ddag} \and
Yang You\inst{1}\textsuperscript{\ddag}
}

\authorrunning{Y. Liu et al.}

\institute{National University of Singapore \and
OpenGVLab, Shanghai AI Laboratory \and
The University of Hong Kong}
\maketitle
\def\thefootnote{*}\footnotetext{equal contribution}\def\thefootnote{\arabic{footnote}}
\def\thefootnote{\dag}\footnotetext{project lead}\def\thefootnote{\arabic{footnote}}
\def\thefootnote{\ddag}\footnotetext{corresponding author}\def\thefootnote{\arabic{footnote}}
\begin{abstract}
Visual-language pre-training has achieved remarkable success in many multi-modal tasks, largely attributed to the availability of large-scale image-text datasets.
In this work, we demonstrate that Multi-modal Large Language Models (MLLMs) can enhance visual-language representation learning by establishing richer image-text associations for image-text datasets.
Our approach is simple, utilizing MLLMs to extend multiple diverse captions for each image.
To prevent the bias introduced by MLLMs' hallucinations and monotonous language styles, we propose \textit{``text shearing"} to maintain the quality and availability of extended captions.
In image-text retrieval, without introducing additional training cost, our method consistently obtains 5.6 $\sim$ 35.0\% and 16.8 $\sim$ 46.1\% improvement on Recall@1 under the fine-tuning and zero-shot settings, respectively.
Notably, we obtain zero-shot results that are comparable to fine-tuning on target datasets, which encourages more exploration of the versatile use of MLLMs. The datasets and codes are available at: \href{https://github.com/lyq312318224/MLLMs-Augmented}{https://github.com/lyq312318224/MLLMs-Augmented}.
\end{abstract}
\input{sec/1_intro}

\input{sec/2_related_work}
\input{sec/3_method}
\input{sec/4_experiments}

\input{sec/5_conclusion}

\bibliographystyle{splncs04}
\bibliography{main}
\clearpage
\input{sec/supp}
\end{document}

%% file: sec/1_intro.tex
\section{Introduction}
\label{sec:intro}

\begin{figure}[htbp]
\centering
\begin{subfigure}{0.25\textwidth}
   \includegraphics[width=\textwidth]{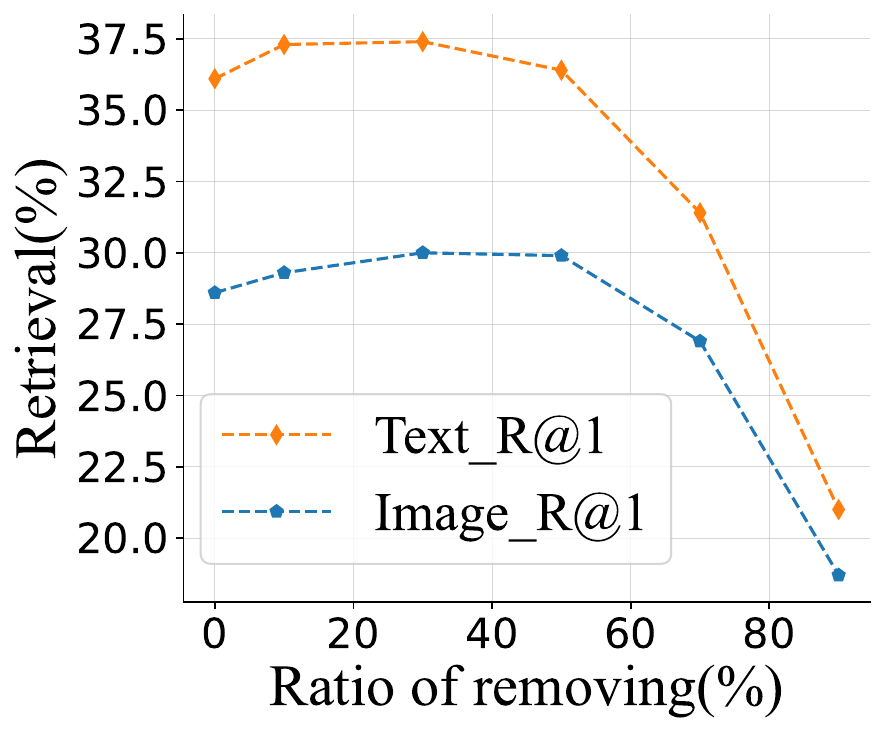}
    \caption{Data Removing Ratios vs. Performance}
    \label{fig:datascale} 
    \includegraphics[width=\textwidth]{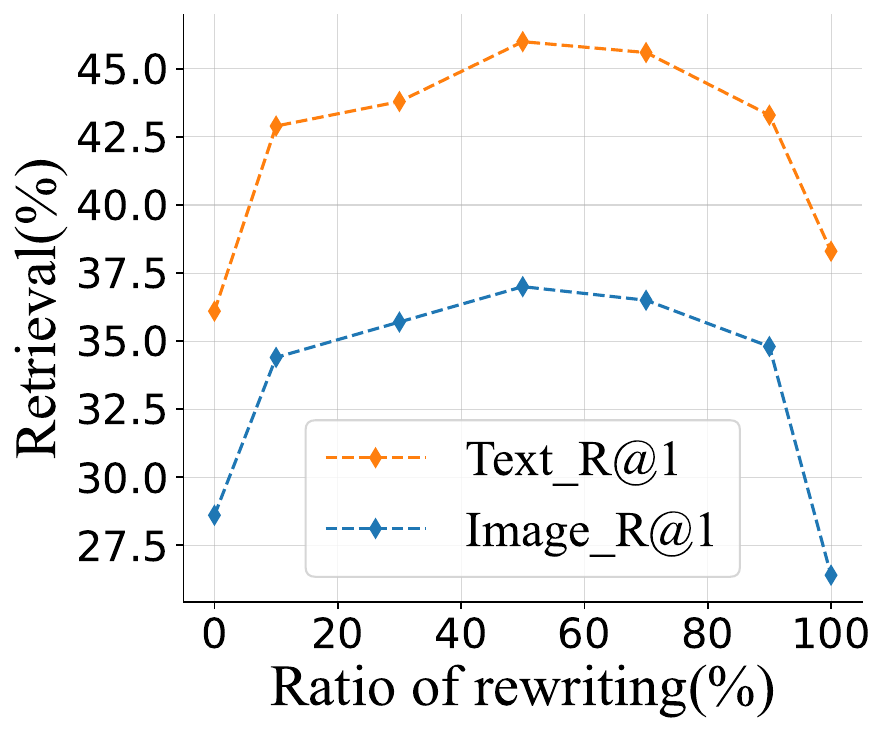}
    \caption{Data Rewriting Ratios vs. Performance}
    \label{fig:caption_match} 
\end{subfigure}
\hspace{0.05\textwidth}
\begin{subfigure}{0.6\textwidth}
    \includegraphics[width=\textwidth]{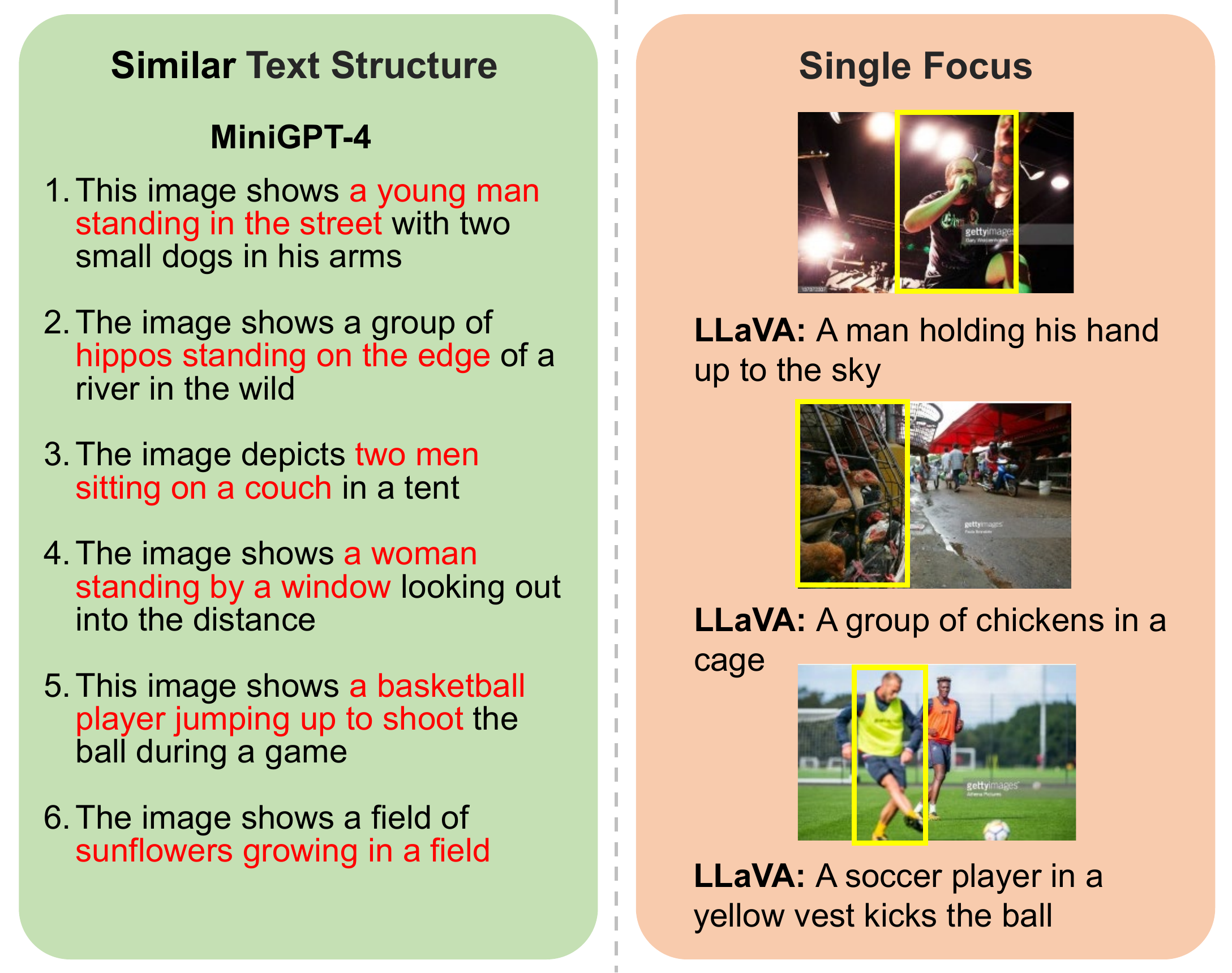}
    \caption{The analysis of captioning the image using MiniGPT-4 and LLaVA.}
    \label{fig:overview}
\end{subfigure}
\caption{The performance in (a) and (b) is zero-shot image-text retrieval results on MSCOCO using pretrained BLIP. (c) shows that using a single model to generate captions can easily result in similar text structure and insufficient global considerations.}
\label{fig:intro_motivation}
\end{figure}

Visual-language pre-training has achieved remarkable success in image-text retrieval\cite{jia2021scaling,kim2021vilt}, image classification\cite{radford2021learning,yu2022coca}, vision question answering~\cite{alayrac2022flamingo,li2022blip}, and image caption generation~\cite{li2023blip,yu2022coca}. This success can be attributed to the large-scale datasets collected from the Internet, such as CC3M~\cite{sharma2018conceptual}, CC12M~\cite{changpinyo2021conceptual}, YFCC100M~\cite{thomee2016yfcc100m}, LAION400M~\cite{schuhmann2021laion}, etc.
However, most of these datasets include a non-negligible portion of noisy and mismatched image-text pairs~\cite{jia2021scaling,wang2023too,lai2023scarcity}, which largely affects the visual-language representation learning. 
One of the most straightforward approaches is: utilizing pre-trained models to recognize and remove mismatched pairs based on heuristic rules~\cite{fan2023improving,abbas2023semdedup,cao2023less,maini2023t}.

These methods indeed reduce the influence of mismatched pairs. However, simply removing mismatched pairs leads to a serious problem: \textit{the number of training pairs is also reduced.}
As illustrated in Figure~\ref{fig:datascale}, the performance of image-text retrieval consistently drops a lot when the removal ratio is large.
Most recently, several works~\cite{fan2023improving,zhu2023chatgpt,nguyen2023improving} demonstrate that LLM and MLLM can be used as re-writers to improve caption quality without reducing the number of training pairs. Unfortunately, these re-writers inevitably introduce their caption styles, \textit{i.e.} text structure, which might disrupt the distribution of the original captions and lead to difficulties in learning better visual-language representations.

To investigate the characteristics of previous rewriting works~\cite{fan2023improving,zhu2023chatgpt,nguyen2023improving}, we conduct experiments to evaluate their influences in Figure~\ref{fig:caption_match}. Experimental results empirically confirm that applying MLLMs to enhance the quality of visual-language datasets is a promising approach, \textit{i.e.} improving the original performance largely. Nevertheless, excessive re-writing leads to non-trivial performance drops. Therefore, it is valuable to figure out what led to these drops and verify whether there exists bias from synthetic captions.

We explore the underlying reasons for the bias by analyzing the generated captions' text structures, attention focus, and word frequency. We utilize MiniGPT-4 and LLaVA to caption each image from CC3M, respectively. Here we use the same question \{Describe the $\left \langle image \right \rangle$ in English:\} to prompt the model and set the same maximum number of generated tokens. As shown in Figure~\ref{fig:overview} and Figure~\ref{fig:unique_styles}, we have the following observations. 1). MLLMs indeed have their inherent text structures and focus. 2). Applying a single MLLM makes it hard to provide comprehensive captions. 3). There are great diversities between the word frequency statistics of different MLLMs and raw captions.

Considering the limitations of a single model in capturing diverse image captions, we employ multiple MLLMs to enrich visual-language associations from different perspectives, thereby improving visual-language representation learning during pre-training.
To improve the quality and availability of synthetic captions, the following specific designs are proposed. First, we crop the captions generated by MLLMs to match the length of the original captions, which is called ``text shearing''. 
This not only reduces the repetitive occurrence of common words in synthetic captions, alleviating the caption collapse problem identified in~\cite{wang2023too}, but also preserves the semantic concepts closest to the image, thereby mitigating the impact of hallucinations typically appearing in the later part of the generated text, as verified in~\cite{zhou2023analyzing}.
Second, to obtain a set of comprehensive captions for each image, we keep the raw and extended captions from MLLMs simultaneously for standard visual-language pre-training.

Our approach exhibits the following characteristics: 1). It is compatible with multiple visual-language pre-training frameworks like CLIP~\cite{radford2021learning} and BLIP~\cite{li2022blip}, demonstrating significant performance improvements across various downstream tasks without introducing additional training overhead. For example, in MSCOCO and Flickr30K's zero-shot image-text retrieval, our method obtains 16.8$\sim$ 46.1\% Recall@1 improvement. In zero-shot image classification, our method achieves an average performance improvement of 13.4 on 15 common classification datasets and 13.1 on ImageNet~\cite{deng2009imagenet}. 2). For image-text retrieval, our zero-shot CLIP outperforms the vanilla CLIP fine-tuned on the MSCOCO and Flickr30K respectively by 9.9\% and 23.5\%. (The above results are all based on pre-training on CC3M.)
3). When scaling to large datasets like CC12M and YFCC15M, our method continues to deliver substantial performance improvements.

%% file: sec/2_related_work.tex
\section{Related Work}
\subsection{Improving image-text datasets}
Numerous studies~\cite{fang2022data,nguyen2022quality,gadre2023datacomp} have emphasized the significance of high-quality image-text datasets in influencing the transfer performance of visual-language pre-training in downstream tasks. 
SemDeDup~\cite{abbas2023semdedup} enhances data efficiency and out-of-distribution performance by identifying and removing semantically duplicated data pairs at the embedding level. 
Cao et al.~\cite{cao2023less} introduced an approach to improve dataset quality by removing samples that contain text in images. 
T-MARS~\cite{maini2023t} masks the text region in the image and then filters out samples with low CLIP~\cite{radford2021learning} similarity.
These methods inevitably lose a lot of visual information when filtering out samples, so some methods try to obtain higher-quality data by rewriting captions. 
Santurkar et al.~\cite{santurkar2022caption} underscored the importance of synthetic captions and employed pre-trained language models to augment textual content. 
Gadre et al.~\cite{gadre2023datacomp} introduced the DataComp benchmark for multi-modal datasets.
Fan et al.~\cite{fan2023improving} leveraged the in-context learning capacity of large language models to rewrite captions, enriching the language structure while preserving core semantics.
Zhu et al.~\cite{zhu2023chatgpt} employed ChatGPT~\cite{openai2022chatgpt} and BLIP-2~\cite{li2023blip} interactively to generate captions with rich visual information.
Nguyen et al.~\cite{nguyen2023improving} used BLIP-2~\cite{li2023blip} to rewrite captions for low-matching image-text pairs and combined original and generated captions for training.
Lai et al.~\cite{lai2023scarcity} proposed enhancing visual information in captions by fusing the original caption with captions generated by LLaVA~\cite{liu2023llava}, leveraging large language models.

While approaches based on caption rewriting have yielded promising results, relying on a single model for rewriting introduces the model's inherent bias and poses challenges in establishing a unified representation between vision and language. In contrast, our method introduces more accurate and diverse descriptions for a single image while retaining rich visual information.
\subsection{Multimodal large language models}
Existing Multimodal Large Language Models (MLLMs) primarily rely on three key technologies: Multimodal Instruction Tuning (M-IT), Multimodal In-Context Learning (M-ICL), and Multimodal Chain-of-Thought (M-CoT)~\cite{yin2023survey}.
M-IT facilitates strong transfer performance by fine-tuning the model on datasets with specific instruction formats. Notable models employing this technology include LLaMA-Adapter~\cite{zhang2023llamaadapter,gao2023llamaadapterv2}, LLaVA~\cite{liu2023llava,liu2023improved}, MiniGPT-4~\cite{zhu2023minigpt}, InstructBLIP~\cite{instructblip}, Qwen-VL~\cite{bai2023qwen}, and NExT-GPT~\cite{wu2023nextgpt}.
M-ICL is a type of analogy learning from a limited number of samples. Models like Flamingo~\cite{alayrac2022flamingo}, Otter~\cite{li2023otter}, and HuggingGPT~\cite{shen2023hugginggpt} are developed using this approach.
M-CoT necessitates models to not only provide answers but also reasoning processes. Representative models include Multimodal-CoT~\cite{zhang2023multicot} and Visual ChatGPT~\cite{wu2023visual}.
These three technologies are not mutually exclusive, and many models effectively combine multiple technologies. 
Since these MLLMs are often trained on the billion-level dataset, they have extremely rich knowledge and excellent visual understanding and expression capabilities, which can be used to caption the images.

%% file: sec/3_method.tex
\begin{figure*}[t]
    \centering
    \includegraphics[width=\textwidth]{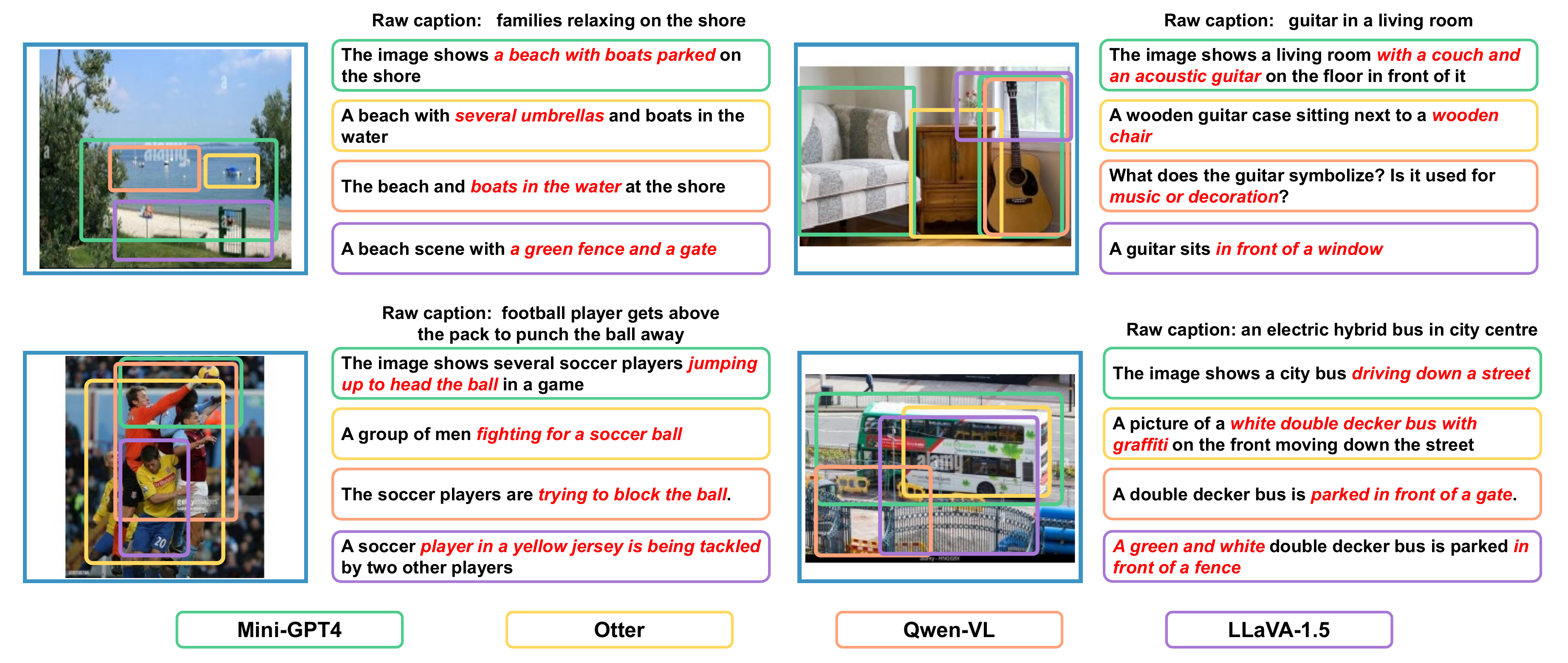}
    \caption{The illustration of our method. Different MLLMs jointly construct one-to-many image-text pairs with richer semantic associations.}
    \label{fig:illustration}
\end{figure*}
\section{Method}
\label{sec:method}
\subsection{Preliminaries}
In this section, we briefly introduce the preliminaries of visual-language pre-training. CLIP~\cite{radford2021learning} is a classic work that uses image-text pairs for contrastive learning. It employs an encoder-only architecture, optimizing the encoder through a contrastive loss on image-text pairs. During training, when a batch of $N$ image-text pairs \{$x_I$ , $x_T$ \} is sampled, the contrastive loss for an image can be defined as follows:
\begin{equation}
L_I = \mathrm{clip\textunderscore contrast}(x_I,x_T)
\label{eq:clip}
\end{equation}
Where $\mathrm{clip\textunderscore contrast}$ is image-text contrastive loss introduced in ~\cite{radford2021learning}. The loss for text is computed in the same manner and can be denoted as $L_T$. The total loss for training is $L=(L_I+L_T)/2$. 

BLIP~\cite{li2022blip} is an encoder-decoder based visual-language pre-training architecture. Its three main pre-training objectives (image-text contrastive loss (ITC), image-text matching loss (ITM), and language modeling loss (LM)) are jointly optimized during the training process.
Specifically, ITC is similar to that in CLIP. ITM aims to capture the fine-grained alignment between vision and language. LM is used to optimize the decoder.

As highlighted in many works~\cite{yu2022coca,mu2022slip,li2023scaling}, visual-language pre-training heavily depends on extensive image-text datasets. The presence of low-quality image-text data can significantly compromise the model's performance. Therefore, improving the multi-modal dataset has become an expected direction.

\subsection{Overview}
In this work, we employ multiple advanced MLLMs to augment the visual-language representation learning from a data-centric perspective. We demonstrate that different MLLMs can generate accurate and diverse captions in Figure~\ref{fig:illustration}. The goal of this approach is to utilize the MLLMs to establish richer image-text associations in the current datasets. 
Our approach consists of two processes: multi-view caption extractor and text shearing. Specifically, for each image from an image-text dataset, we first introduce multiple advanced MLLMs to synthesize extended captions. Then, by comprehensively analyzing the characteristics of these extended captions, we have a key observation: MLLMs have their caption styles and might generate hallucinated content. Based on this observation, we propose text shearing to maintain the quality and availability of extended captions. After these processes, the raw and extended captions with corresponding images are jointly used for standard visual-language pre-training.

\subsection{Multi-View caption extractor}
Given a dataset $\mathcal{T}=\{(\bm{x}_I^i,x_T^i)\}_{i=1}^{N}$, containing $N$ paired images $\bm{x}_I^\cdot$ and texts $x_T^\cdot$. We define a model pool $G=\{g_1\cdots g_k\cdots g_K\}$ that includes $K$ advanced MLLMs. For each image $\bm{x}_I^i$,  $G$ is used to obtain rich captions. The operation can be formulated as follows,
\begin{equation}
    C^i = \{c_1^i\cdots c_n^i\cdots c_N^i\}=G(x_I^i,Q)
\label{eq:multicaption}
\end{equation}
Where $C^i$ represents the extended captions for image $x_I^i$, and $Q$ denotes the question input to the MLLM.
For different models in $G$, we use the same simple question template: \{Describe the $\left \langle image \right \rangle$ in English:\} to query the captions.
The simple question has little impact on the diversity of answers, so we can obtain comprehensive captions of each image.

\begin{figure*}[ht]
    \centering
    \includegraphics[width=\textwidth]{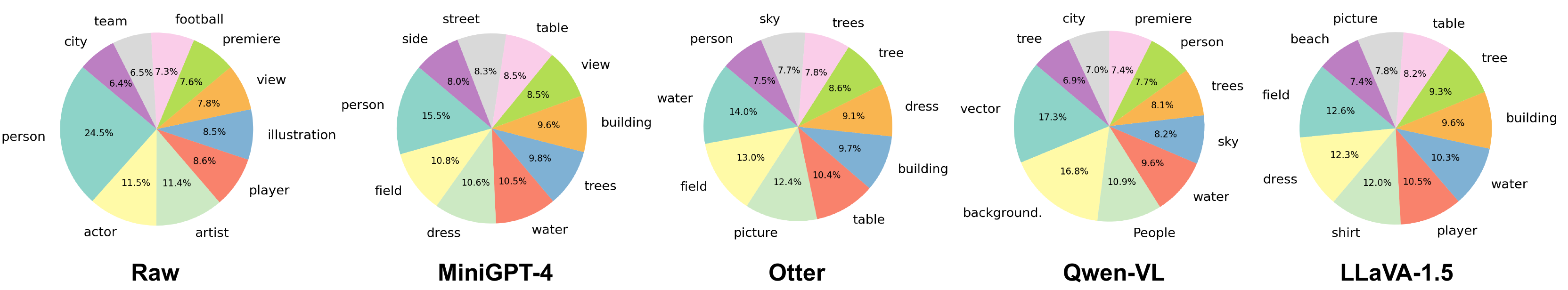}
    \caption{Statistics of common nouns in captions generated by MLLMs. }
    \label{fig:unique_styles}
\end{figure*}

\textbf{How to use these extended captions?}
After obtaining these extended captions, one of the most straightforward ideas is to replace a set of raw captions with new captions for training. Its effectiveness has been evaluated in many previous works~\cite{zhu2023chatgpt,nguyen2023improving}. However, we have a concern about the operation in terms of the difference between raw and new captions. To investigate this, we count the frequency of the most occurring nouns of the extended captions (extracted from images in CC3M by MiniGPT-4, Otter, Qwen-VL, and LLaVA-1.5, examples are shown in Figure~\ref{fig:illustration}) and present the results in Figure~\ref{fig:unique_styles}. One can easily find that different MLLMs output different common words in the generated captions. These diverse words can greatly expand the linguistic concepts of the dataset.

\vspace{-5pt}

\begin{figure}[h]
    \centering
    \begin{subfigure}{0.48\textwidth}
    \includegraphics[width=0.9\textwidth]{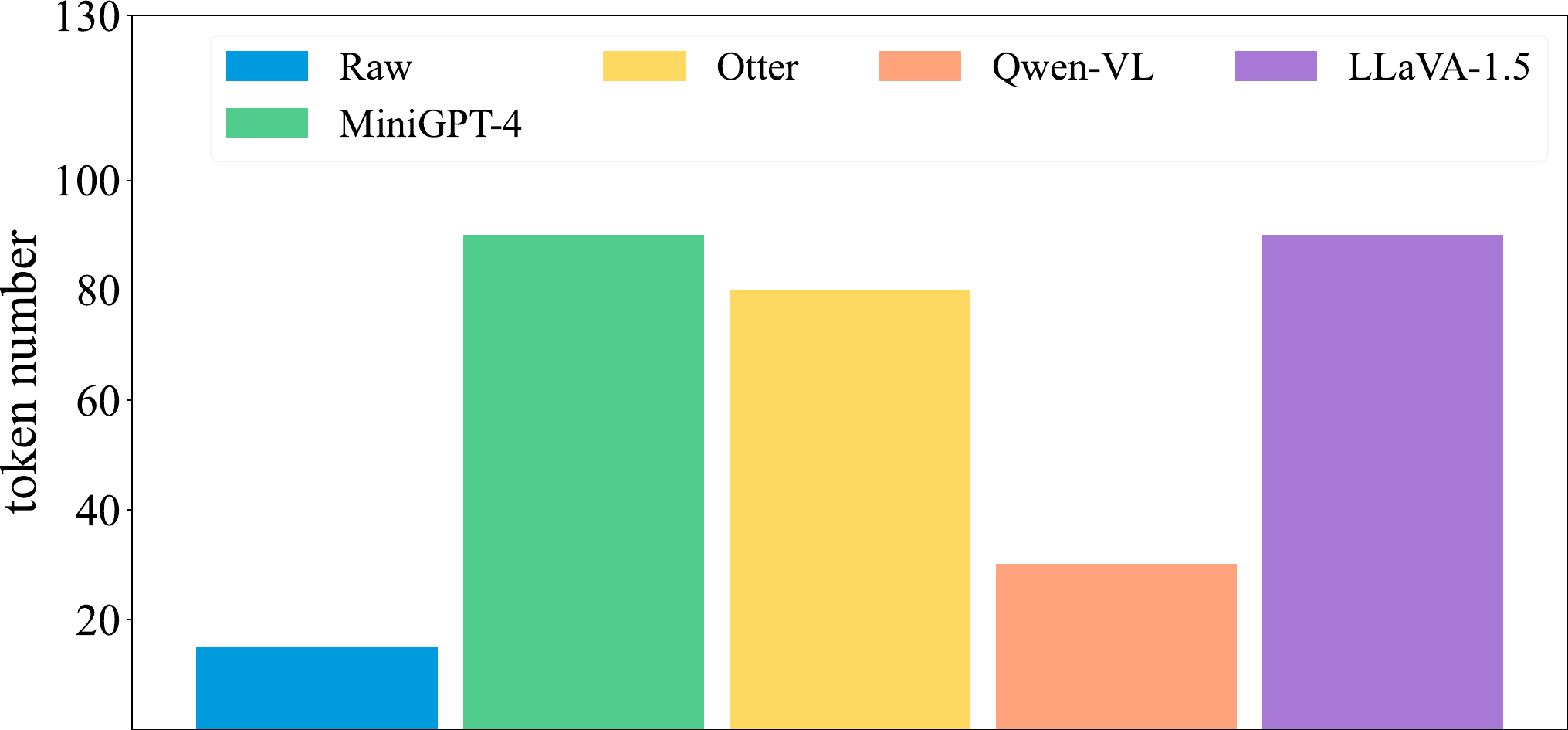}
    \caption{Average length of captions generated using different MLLMs.}
    \label{fig:length}
    \end{subfigure}
    \begin{subfigure}{0.48\textwidth}
    \includegraphics[width=\textwidth]{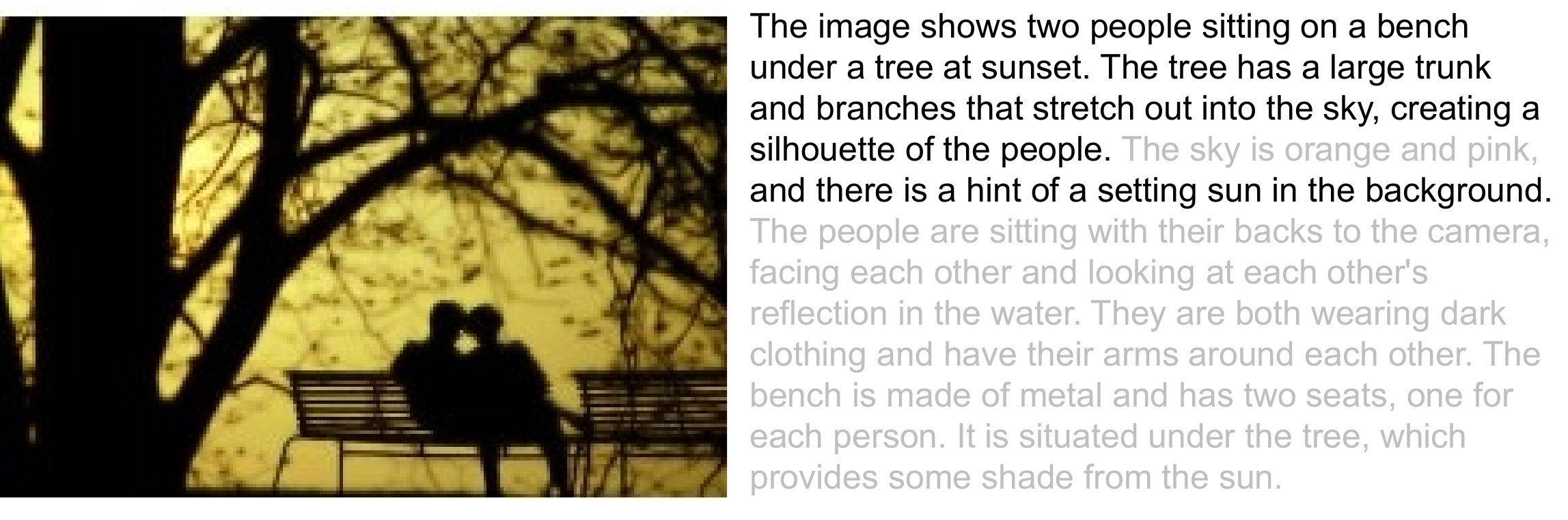}
    \caption{Raw caption:$\left \langle \text{lovers on a park bench.} \right \rangle$ Here we use MiniGPT-4 to caption the image and mark the hallucinations of the generated caption in \textcolor{gray}{gray}.}
    \label{fig:hallucinations}
    \end{subfigure}
    \caption{Analysis of extended captions.}
\end{figure}
\vspace{-5pt}

Based on the above analysis, we plan to use all the extended captions for training to improve the diversity of captions.
However, there is another concern about the difference in the length of the raw and extended captions.   
We present the average length comparisons of these captions in Figure~\ref{fig:length}. These captions are generated by MLLMs under default settings.%
It can be seen that the generated captions by MiniGPT-4 and LLaVA-1.5 are notably longer, \textit{i.e. }90 tokens. While raw caption is shorter, \textit{i.e. }15 tokens. To investigate the influence of varying caption lengths on pre-training, we visualize the longer caption generated by MiniGPT-4 in Figure~\ref{fig:hallucinations}. We find that longer captions indeed produce hallucinations that do not match the image, and these hallucinations often appear in the later part of the generated text, which has an impact on representation learning. The same conclusion is also verified in ~\cite{zhou2023analyzing}. It states that as generation progresses, the accumulation of past hallucinatory information and uncertainty can derail the model and lead to new hallucinations.

\subsection{Text Shearing}
To eliminate the impact of excessive caption length, we propose text shearing to make the length of the extended captions as same as the raw ones. Specifically, we add a token number $T$ as a limit when using MLLMs to generate captions. 
To prevent the generated captions from having incomplete expressions, we intercept the first complete clause in the generated caption.
Therefore, Eq.~\ref{eq:multicaption} can be rewritten as:
\begin{equation}
    C_i' = \{c_1^{i'}\cdots c_n^{i'}\cdots c_N^{i'}\}=G(x_I^i,Q,T)
\label{eq:control_length}
\end{equation}
For the setting of $T$, we use a simple but effective strategy. We set it to the average length of the original captions to mimic human annotation. By controlling the length of generated captions, we hope to effectively reduce the impact of MLLMs' hallucinations and their monotonous language styles.

Then, we jointly use original captions and newly generated captions to construct a new dataset. The new image-text pairs can be expressed as $\{x_I , x_T'\}=\{x_I , c_1^{i'}\},\{x_I , c_2^{i'}\}\cdots,\{x_I , c_k^{i'}\}$, where $k$ represents the number of MLLMs. We use the enhanced dataset for subsequent standard visual-language pre-training. 
Taking CLIP as an example, the loss on the image can be written as:
\begin{equation}
L_I = \mathrm{clip\textunderscore contrast}(x_I,x_T')
\end{equation}
The total loss $L=(L_I+L_T)/2$. The parameters $\beta$ for CLIP is updated by minimizing $L$:
\begin{equation}
\beta \leftarrow \argmin _\beta L.
\end{equation}

%% file: sec/4_experiments.tex
\section{Experiments}
\label{sec:experiments}
\subsection{Pre-training Details}
Our method is implemented in Pytorch~\cite{paszke2019pytorch} and trained on a node equipped with 8 NVIDIA A100 GPUs. For visual language pre-training, we follow the implementation of CLIP~\cite{radford2021learning} and BLIP~\cite{li2022blip} respectively.
For CLIP, we train the model based on the open source code OpenCLIP~\cite{ilharco_gabriel_2021_5143773}. During training, our batch size is set to 320. The number of epochs is set to 6 to better align the training overhead. The model architecture is set as VIT-B-16. Other parameters are the same as the default values in OpenCLIP~\cite{ilharco_gabriel_2021_5143773}. 
For BLIP, we use a ViT-B-16~\cite{dosovitskiy2020image,touvron2021training} pre-trained on ImageNet~\cite{deng2009imagenet} as the image encoder and $\text{Bert}_\text{base}$~\cite{devlin2018bert} as the text encoder. Regarding image preprocessing, we employ random cropping to achieve $224 \times 224$ during pre-training and increase the image resolution to $384 \times 384$ during fine-tuning. We apply RandomAugment~\cite{cubuk2020randaugment} for image augmentation. For optimization,  we utilize an AdamW~\cite{loshchilov2017decoupled} optimizer with a learning rate warm-up to 3e-4, combined with linear decay at a rate of 0.9. Our training batch size is set to 1280, and the number of training epochs is 4.

Regarding the usage of multi-modal large language models, we employed Mini-GPT4-Vicuna13B~\cite{zhu2023minigpt}, Otter-Image-MPT7B ~\cite{li2023otter}, Qwen-VL-Chat~\cite{bai2023qwen}, and LLaVA-v1.5-13B~\cite{liu2023improved}. During caption generation, we impose a maximum token limit of 30 to control the caption length. For the synthetic caption generation, we use beam search with the number of beams equal to 1.

We use CC3M~\cite{sharma2018conceptual}, CC12M~\cite{changpinyo2021conceptual} and YFCC15M as pre-training datasets. Among them, YFCC15M is a subset of YFCC100M~\cite{thomee2016yfcc100m} that contains English descriptions of images. Due to the inaccessibility of some images, the version of the dataset we used contains fewer images than the original dataset. Among them, on CC3M, compared with the 3.3M version of the original dataset, the version we used contains 2.6M image-text pairs; on CC12M, compared with the 12.4M version of the original dataset, the version we used contains 11.1M image-text pairs; On YFCC15M, compared to the 15M of the original dataset, our version contains 14.8M image-text pairs.

\input{tables/image-text-retrieval}
\subsection{Evaluation}
We evaluate the performance of the multi-modal task in Table~\ref{tab:image-text-retrieval} and Table~\ref{tab:vqa_nlvr}. We also provide results on the image classification task in Table~\ref{table:zeroshot-main} and Table~\ref{table:linear-main}.

\textbf{Image-Text Retrieval.}
Image-text retrieval serves as a crucial metric for evaluating the bridging capability between different modalities in a model. We pre-train CLIP and BLIP on CC3M, conducting image-to-text retrieval (TR) and text-to-image retrieval (IR) testing on the MSCOCO~\cite{lin2014microsoft} and Flickr30K~\cite{plummer2015flickr30k} datasets. Results in Table~\ref{tab:image-text-retrieval} demonstrate significant performance improvements in both zero-shot and fine-tuning retrievals for both CLIP and BLIP. Notably, our approach enables CLIP's zero-shot retrieval performance to outperform models finetuned on the target dataset. Specifically, when using CLIP for zero-shot retrieval on MSCOCO, the R@1 of TR and IR increase by 27.2 and 19.4, respectively. For zero-shot retrieval on Flickr30K, the R@1 of TR and IR also improves by 46.1 and 35.4, respectively. Similarly, our method also achieves an improvement of 16.8$\sim$23.4 when using BLIP for zero-shot retrieval. These significant performance improvements indicate that the pre-trained model learns better image-text representation.

\input{tables/clip_zeroshot}
\textbf{Image Classification.}
We evaluate the performance of our method on the task of image classification. In Table~\ref{table:zeroshot-main}, we pre-train a CLIP with a VIT-B-16 architecture on CC3M and CC12M respectively, followed by testing on sixteen common image classification datasets. We compare our results with ~\cite{fan2023improving}.
Our method exhibits significant performance improvements across most datasets, as demonstrated. For the CLIP pre-trained on CC3M, there's an average improvement of 13.4 on 15 datasets and a 13.1 improvement on ImageNet~\cite{deng2009imagenet}. For the CLIP pre-trained on CC12M, there's an average improvement of 11.1 and 10.2 respectively. These results highlight the effective enhancement of representation learning achieved by our method.

\input{tables/linear_probing}
\textbf{Linear-Probing.}
We also explore the linear-probing performance in image classification. We use CLIP pre-trained on CC3M and CC12M for evaluation respectively. Similar to the zero-shot image classification, we compare our method with~\cite{fan2023improving} across 15 common datasets. The results in Table~\ref{table:linear-main} show that our method also has certain improvements for linear probing. This shows that rich text concepts contribute to the effective training of visual encoders. As the visual encoder learns each image, it benefits from supervision through multiple accurate and diverse captions. This allows it to efficiently acquire a universal representation of the image, leading to a more effective characterization of the image and text embedding space.

\textbf{Vision Question Answering.}
Visual Question Answering (VQA)~\cite{antol2015vqa} is the task of providing answers based on an image and a question. We conduct tests on the VQAv2~\cite{goyal2017making}, A-OKVQA~\cite{schwenk2022okvqa} and OK-VQA~\cite{marino2019ok} using BLIP pre-trained on CC3M, CC12M and YFCC15M, respectively. The results are presented in Table~\ref{tab:vqa_nlvr}. Similar to BLIP~\cite{li2022blip}, we also consider VQA as an answer-generation task. The consistent performance improvements obtained through our method indicate that the model has acquired a more robust visual-language representation from datasets enriched with MLLMs' knowledge. Besides, the improvement on A-OKVQA and OK-VQA indicates that the model has more common sense and world knowledge. These results suggest an enhanced capability in visual understanding and language capabilities together.

\input{tables/cc3m_othertasks}
\textbf{Visual Reasoning.}
In the Natural Language Visual Reasoning (NLVR$^2$) task~\cite{suhr2018corpus}, the model is required to perform multi-modal reasoning by analyzing two images and a natural language question. We conduct an evaluation using the pre-trained BLIP on CC3M, CC12M, and YFCC15M, respectively. As depicted in Table~\ref{tab:vqa_nlvr}, our method consistently delivers improved performance. This shows that the model makes certain progress in natural language understanding, visual recognition, and logical reasoning.

\textbf{Image Captioning.}
Image Captioning is the task of generating a text description of the image content given an image.
We conduct tests on COCO and Nocaps~\cite{agrawal2019nocaps} using the BLIP pre-trained on CC3M, CC12M, and YFCC15M, respectively.
Following the approach in~\cite{li2022blip}, we initially fine-tune the pre-trained model with the LM loss on COCO.
To achieve better results, we also add "a picture of" at the beginning of the prompt.
The results are presented in Table~\ref{tab:vqa_nlvr}, indicating improvements in BLEU@4 and CIDEr metrics.
These findings suggest that our method enhances the model's understanding of the relationship between texts and images, resulting in higher similarity with human annotations during the captioning process.

\begin{wraptable}{r}{0.5\textwidth} %
  \centering
\footnotesize
\resizebox{0.5\textwidth}{!}{
		\begin{tabular}	{c|llll}
		Dataset &  R@1$\uparrow$ &  R@5$\uparrow$ &  R@10$\uparrow$ & MdR$\downarrow$\\
		\shline
		       CC3M~\cite{sharma2018conceptual} & 26.0&46.3&58.0& 7.0\\
         Ours & 28.3$_{\textcolor{mygreen}{+2.3}}$&50.6$_{\textcolor{mygreen}{+4.3}}$&60.7$_{\textcolor{mygreen}{+2.7}}$ &5.0$_{\textcolor{mygreen}{-2.0}}$ \\ 
        \hline
        CC12M~\cite{changpinyo2021conceptual}& 34.0 &56.7 & 68.5 & 4.0 \\
         Ours &36.2$_{\textcolor{mygreen}{+2.2}}$&60.6$_{\textcolor{mygreen}{+3.9}}$&70.6$_{\textcolor{mygreen}{+2.1}}$ & 3.0$_{\textcolor{mygreen}{-1.0}}$\\
        \hline
        YFCC15M~\cite{thomee2016yfcc100m}& 28.2 & 48.0 & 63.2 & 6.0\\
         Ours & 32.3$_{\textcolor{mygreen}{+4.1}}$&  53.6$_{\textcolor{mygreen}{+5.6}}$&  65.4$_{\textcolor{mygreen}{+2.2}}$ &  4.0$_{\textcolor{mygreen}{-2.0}}$\\
        \hline
		\end{tabular}
  }
      \caption
	{The retrieval performance on the video-language retrieval dataset MSRVTT. R@K stands for recall at K and MdR stands for median rank.
	}
	\label{tab:zsl_video_retrieval}
\end{wraptable}
\textbf{Video-Language Task.}
Text-to-video retrieval is a metric for evaluating a model's generalization ability in video-language tasks. We evaluate our method on the MSRVTT dataset.
Following the ~\cite{li2022blip}, we fine-tune the model on COCO. For video input, we uniformly sample 8 frames from it to get a sequence.
The results in Table~\ref{tab:vqa_nlvr} demonstrate a stable performance improvement achieved by our method. This suggests that robust visual-language representation learning may be key to video-text retrieval.

\subsection{Analysis}
\textbf{Scaling Ability.} Notably, as shown in Table~\ref{tab:vqa_nlvr}, our method shows certain improvements when pre-training BLIP on CC12M~\cite{changpinyo2021conceptual} and YFCC15M~\cite{thomee2016yfcc100m}. This indicates that our method is scalable to larger datasets to some extent. It also emphasizes the significance of constructing well-represented image-text pairs for enhancing visual-language pre-training.

\textbf{Training Cost.} To ensure a fair comparison with baseline methods, we carefully craft the training schedule. Considering that the number of image-text pairs in the augmented dataset is $k$ times the number of image-text pairs in the original dataset, we adjust the pre-training epochs to be $1/k$ of the original epochs. By doing this, we avoid introducing additional training overhead.

\begin{figure*}[t]
\centering
\begin{subfigure}{0.24\textwidth}
    \includegraphics[width=\textwidth]{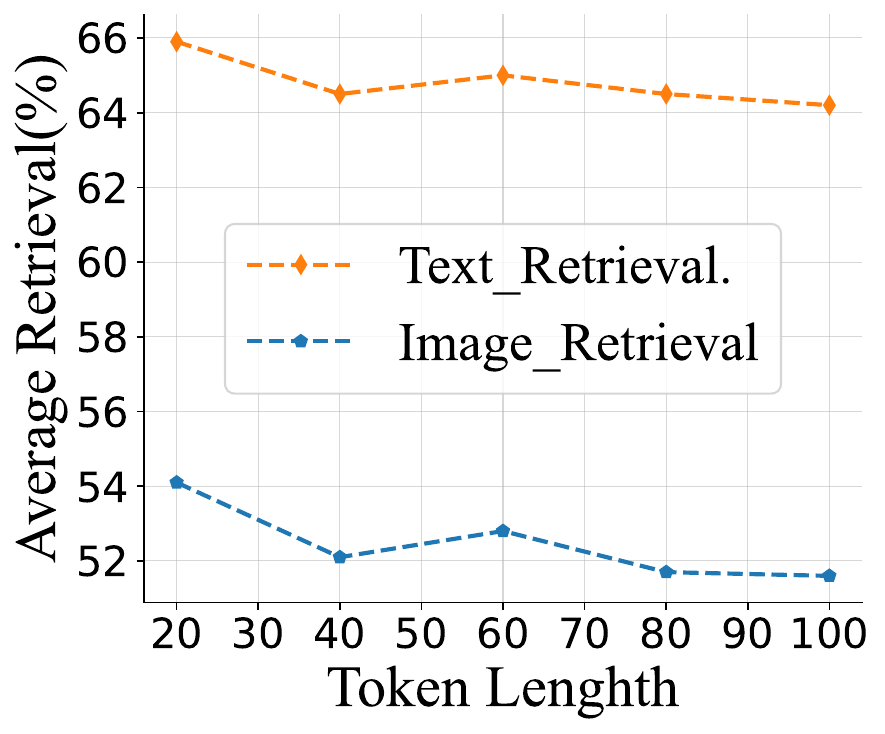}
    \caption{Token Length vs. Performance}
    \label{fig:TokenLength} 
\end{subfigure}
\begin{subfigure}{0.24\textwidth}
    \includegraphics[width=\textwidth]{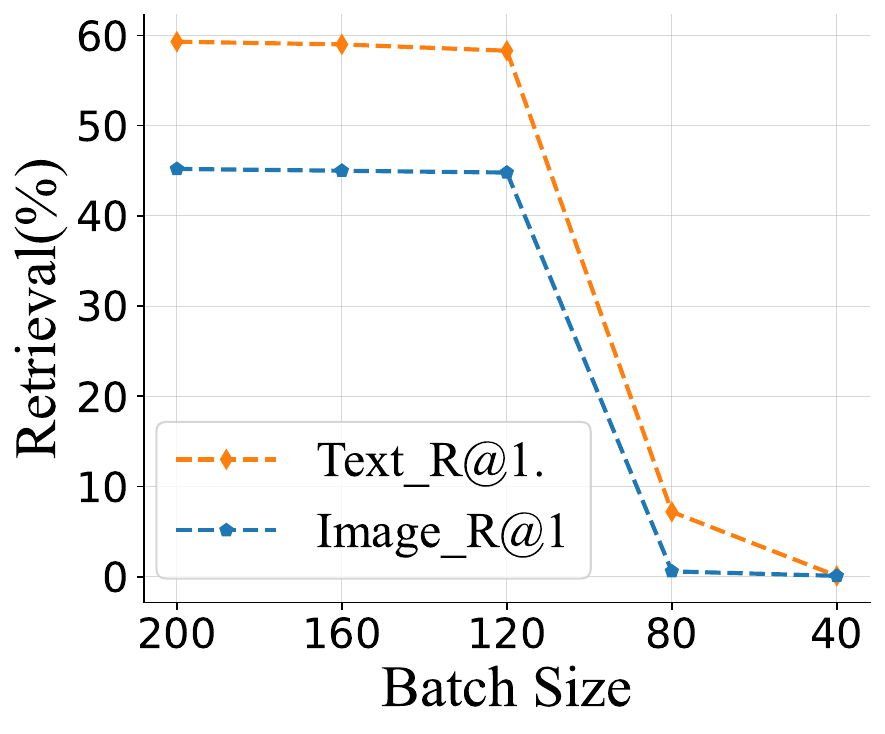}
    \caption{Batch size vs. Performance}
    \label{fig:batchsize} 
\end{subfigure}
\begin{subfigure}{0.24\textwidth}
    \includegraphics[width=\textwidth]{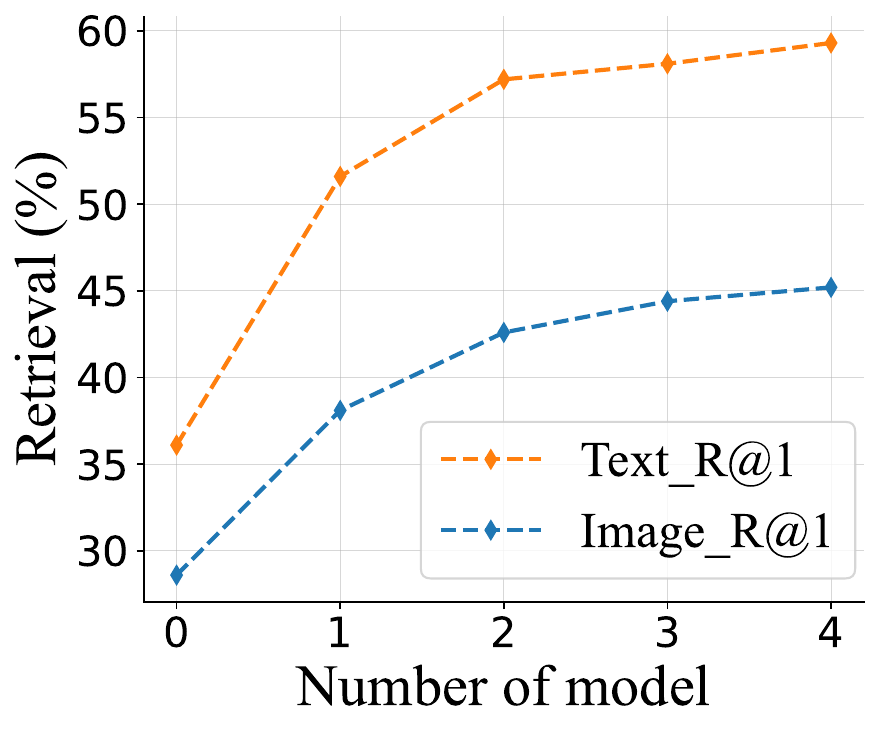}
    \caption{Model Numbers vs. Performance}
    \label{fig:caption_diversity} 
\end{subfigure}
\begin{subfigure}{0.24\textwidth}
    \includegraphics[width=\textwidth]{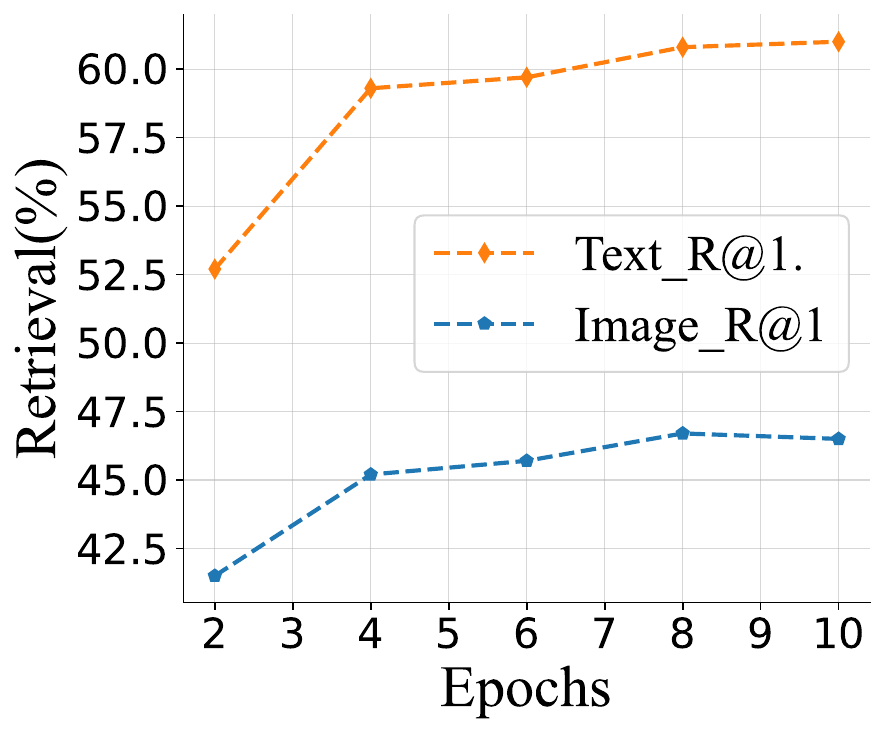}
    \caption{Epochs vs. Performance}
    \label{fig:Epochs} 
\end{subfigure}
\caption{The influence of the token length, batch size, model numbers, and epochs on the performance of visual-language pre-training.}
\label{fig:intro_motivation}
\end{figure*}
\subsection{Ablation Study}
\textbf{Caption Length.} The length of the generated captions is one of the factors affecting visual-language pre-training.
We utilize MiniGPT-4 to generate captions for CC3M with varying max token number limits.
By training BLIP on captions of different lengths and conducting retrieval on MSCOCO, we present the results in Figure\ref{fig:TokenLength} and observe that as the length of the caption increases, the model's performance tends to decrease. This may be because too-long captions
\begin{wrapfigure}{r}{0.5\textwidth} %
  \centering
    \includegraphics[width=0.5\textwidth]{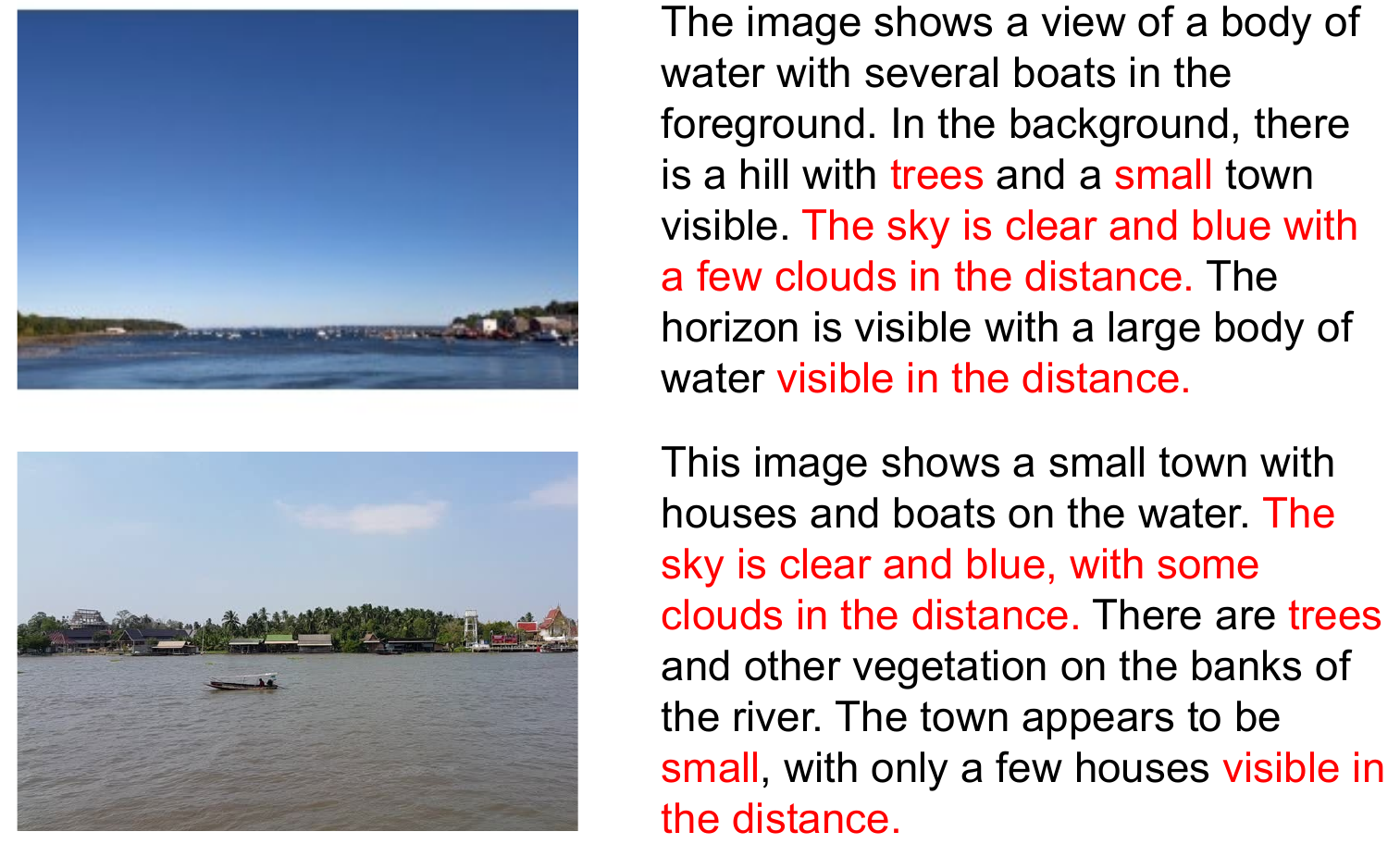}
    \caption{Long captions for different images tend to be similar due to the intrinsic inference manner of the model. Similar content in captions is shown in \textcolor{red}{red}.}
    \label{fig:text_distribution}
\end{wrapfigure}
cause more similar text features between captions of different images.
We illustrate this phenomenon in Figure~\ref{fig:text_distribution}.
When the caption length increases, the generated captions often become identical in many instances. This leads to different images easily mapping to the same text features, creating challenges in learning an effective representation.

\begin{figure*}[t]
    \centering
    \includegraphics[width=1.0\textwidth]{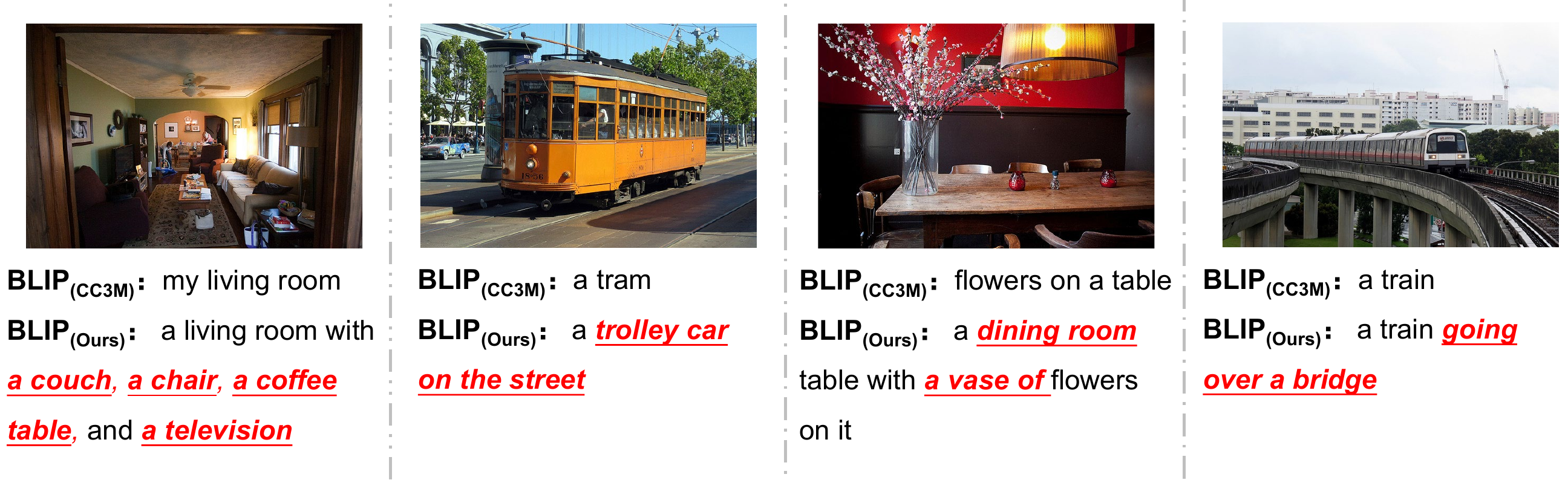}
    \caption{Examples of image captioning using pre-trained BLIP without any fine-tuning. Newly introduced visual concepts are represented in \textcolor{red}{red}. It can be seen that BLIP pre-trained on our dataset has stronger visual concept capture capabilities.}
    \label{fig:Imagecaptioning}
\end{figure*}
\textbf{Batch Size.} We explore the impact of batch size on training in Figure~\ref{fig:batchsize}.
Training models on synthetic captions requires a relatively high batch size compared to using captions annotated by humans.
We attribute this to the fact that, with a small batch size, the main gradient information obtained is noisy, differing significantly from the true gradient.
Conversely, with an increased batch size, the gradient of the sampled distribution becomes more similar to the gradient of the real distribution, enabling the model to generalize better.

\textbf{Number of MLLMs.} We explore the impact of the number of MLLMs on the transfer performance in Figure~\ref{fig:caption_diversity}.
We add four MLLMs\{MiniGPT-4, Otter, Qwen-VL, and LLAVA-1.5\}in order.
Using pre-trained models for zero-shot retrieval on MSCOCO, we observe that, as the number of MLLMs increases, the performance gradually improves.
However, the magnitude of this improvement is gradually decreasing, indicating that the one image's caption information obtained from MLLM is reaching saturation.

\textbf{Training Epochs.} We visualize the curve depicting the number of training epochs versus the model's performance on the retrieval task in Figure~\ref{fig:Epochs}.
It is evident that our method achieves promising performance with only a small number of epochs.
In the main experiment, we choose epoch number 4 to compare with the baseline without increasing the training cost.
When further increasing the number of epochs, our method continues to show certain improvements.

\begin{wraptable}{r}{0.5\textwidth} %
\centering

{
\vspace{-5pt}
\centering
\resizebox{0.5\textwidth}{!}{
\begin{tabular}{c|cccc|c}
    \toprule[0.9pt]
    \multirow{2}{*}{\bf MLLM} & \multicolumn{2}{c}{\bf COCO (R@1)} & \multicolumn{2}{c|}{\bf Flickr30k (R@1)} & \multirow{2}{*}{\bf Avg.} \\
    & I2T & T2I & I2T & T2I & \\
    \hline
    Raw & 36.3 & 28.6 & 62.1 & 51.2 & 44.6 \\
    MiniGPT-4 & 51.6 & 38.1 & 77.0 & 60.5 & 56.8 \\
    Otter & 50.6 & 37.5 & 78.3 & 61.9 & 57.1 \\
    Qwen-VL & 43.4 & 35.4 & 72.1 & 61.7 &  53.2\\
    LLaVA-1.5 & 48.4 & 37.1 & 76.7 & 61.2 & 55.9 \\
    \bottomrule[1.0pt]
  \end{tabular}}
}

\caption{
The retrieval performance of models trained on captions rewritten using one MLLM.}
\label{tab:one_mllm}
\vspace{-15pt}
\end{wraptable}
\textbf{MLLM's Independent Effect.} We also evaluate the performance of rewriting captions using only one MLLM and utilize the generated captions for training.
Captions are rewritten using MiniGPT-4, Otter, Qwen-VL, and LLaVA-1.5 respectively, and the results are presented in Table~\ref{tab:one_mllm}.
The results indicate that there is an upper limit to the performance improvement achievable by rewriting with only one MLLM. Besides, different MLLMs also have different strengths in knowledge.
Combining multiple MLLMs' knowledge proves to be more effective in enhancing visual-language pre-training.

\subsection{Visualization}
\textbf{Image Captioning Visualization.} In Figure~\ref{fig:Imagecaptioning} we illustrate the difference in image captioning between models pre-trained on the original CC3M dataset and our improved dataset.
Without any fine-tuning, our model exhibits a significant improvement in the ability to recognize visual concepts in images.

\begin{figure}[h]
\centering
\begin{subfigure}{0.5\textwidth}
    \includegraphics[width=\textwidth]{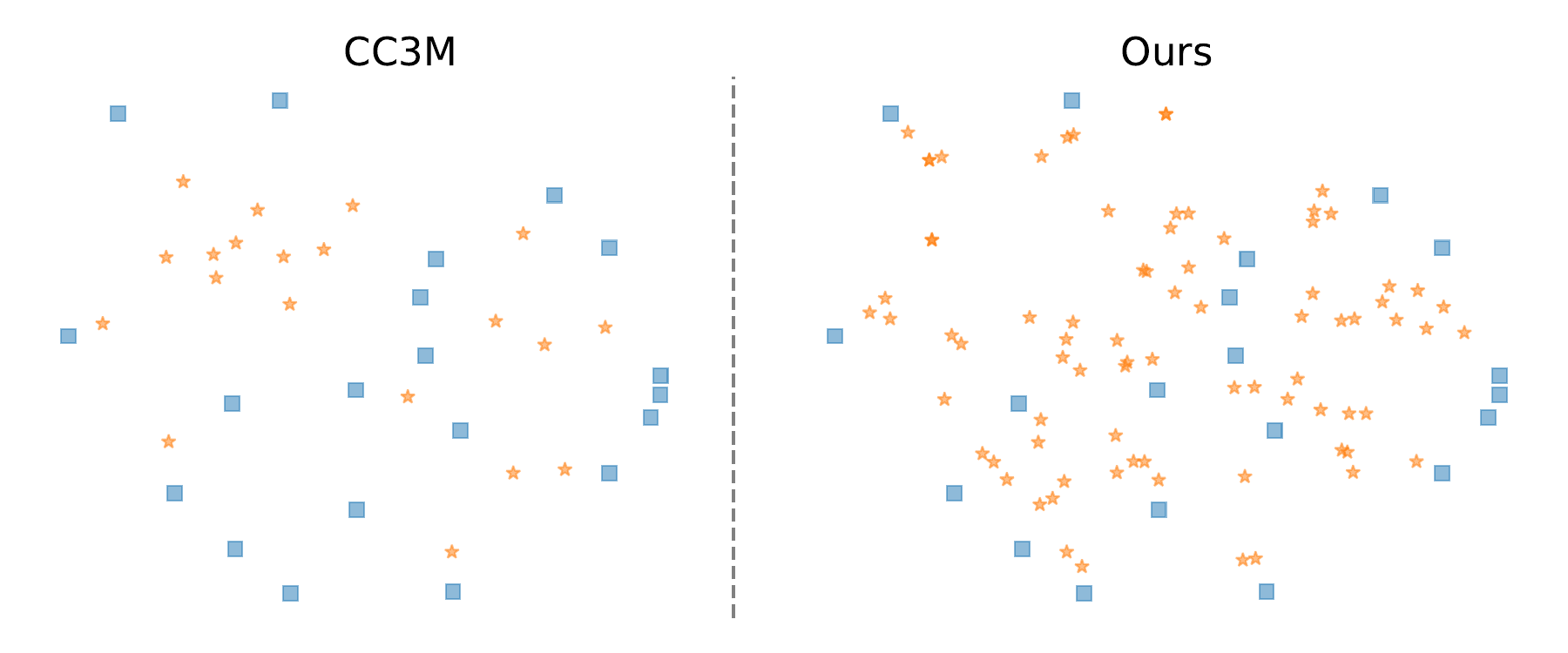}
    \includegraphics[width=0.7\textwidth]{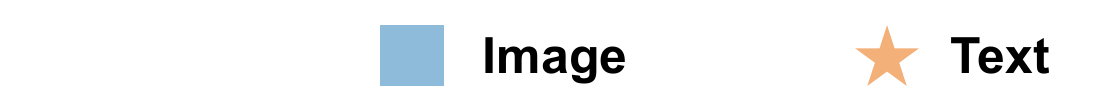}
    \caption{Image-text distribution visualization.}
    \label{fig:itdistribution}
\end{subfigure}
\hspace{0.03\textwidth}
\begin{subfigure}{0.45\textwidth}
    \includegraphics[width=\textwidth]{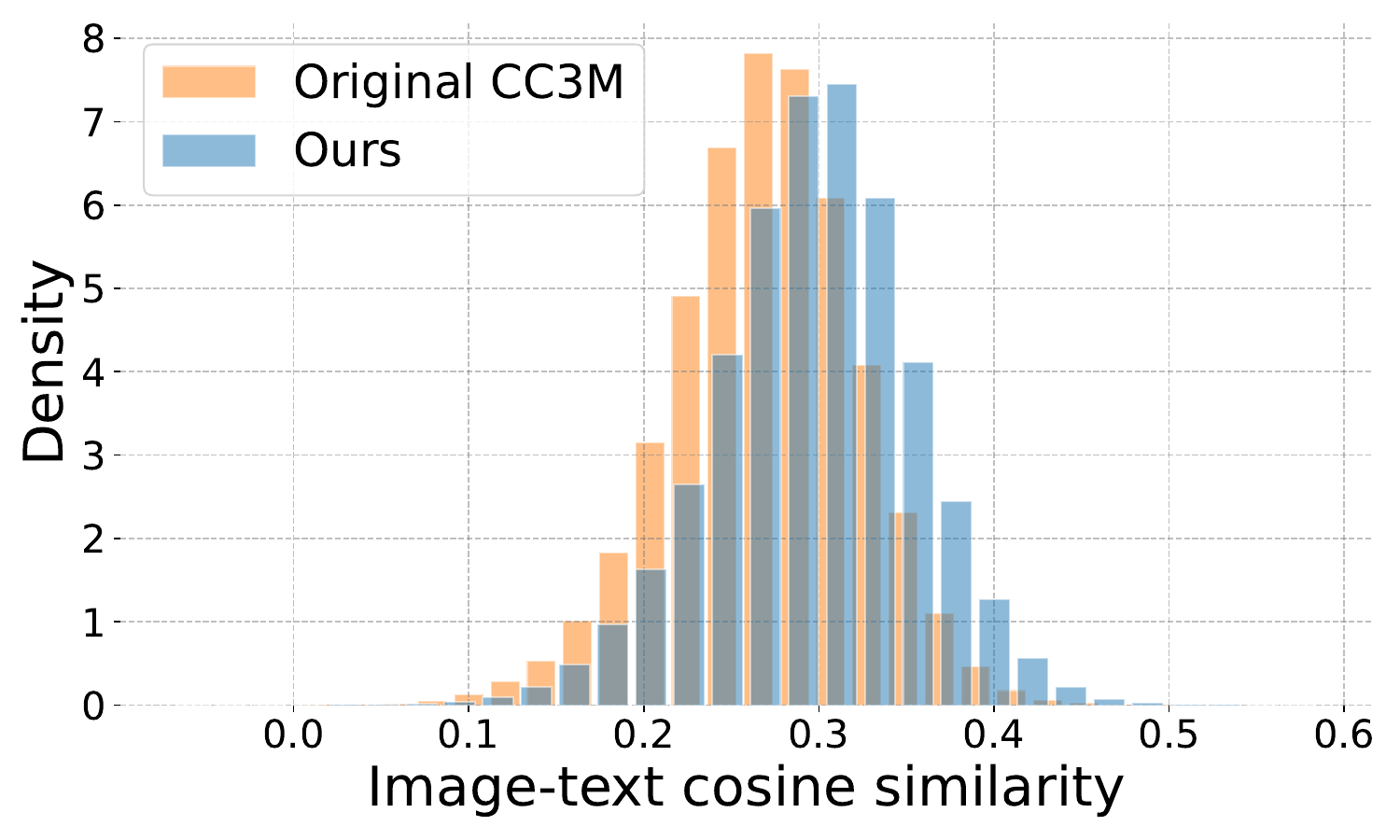}
    \caption{Cosine similarity visualization.}
    \label{fig:cosine} 
\end{subfigure}
\caption{(a) Distribution of images and texts in the original CC3M dataset and our enhanced CC3M dataset. Our method aligns images and texts better in the latent space. (b) Cosine similarity distribution of images and texts. Our method improves the average matching degree of image-text pairs.}
\end{figure}
\textbf{Image-text Distribution Visualization.} We visualize the feature distribution of image-text pairs from the original CC3M and our dataset in Figure~\ref{fig:itdistribution}.
It can be observed that the original CC3M dataset contains numerous discrete and unmatched image-text pairs.
In contrast, our dataset exhibits a distribution where almost all images have corresponding texts.
This distribution perspective explains how our method aligns more continuous images with discrete language, thereby enhancing visual-language representation learning.

\textbf{Cosine Similarity Visualization.} We visualize the cosine similarity comparison between the original CC3M dataset and our enhanced CC3M dataset in Figure~\ref{fig:cosine}. We use a pre-trained CLIP to calculate the cosine distance between image embeddings and text embeddings. Our method exhibits a higher average similarity, which proves the effectiveness in improving the matching degree of the image-text pairs and establishing robust image-text associations for datasets.

\vspace{-5pt}
\subsection{Comparison with More Methods}
\vspace{-5pt}
\input{tables/compared_veclip}
We also compare our work with the contemporaneous work VeCLIP~\cite{lai2023scarcity} in the same setting in Table~\ref{table:VeCLIP}. We use pre-trained CLIP to compare results on image-text retrieval and image classification. It can be seen that our method achieves better performance improvement than the caption fusion used in VeCLIP. This also implies that establishing more continuous image-text correspondences is the key to improving visual-language pre-training on small-scale datasets. 
We also compare our method with ~\cite{nguyen2023improving} in the appendix.

%% file: tables/image-text-retrieval.tex
\begin{table*}[t]
    \centering
    \footnotesize
    \caption{The results of \colorbox{blue!10}{zero-shot image-text retrieval} and fine-tuning image-text retrieval on MSCOCO and Flickr30K.}
    \resizebox{1.0\textwidth}{!}{
    \begin{tabular}{lc|cccccc|cccccc}
    \multirow{3}{*}{Method} & \multirow{3}{*}{Dataset} &
    \multicolumn{6}{c|}{MSCOCO (5K test set)} & \multicolumn{6}{c}{Flickr30K (1K test set)} \\
    & & \multicolumn{3}{c}{Image$\rightarrow$ Text} & \multicolumn{3}{c|}{Text$\rightarrow$ Image} &
    \multicolumn{3}{c}{Image$\rightarrow$ Text} & \multicolumn{3}{c}{Text$\rightarrow$ Image} \\
    & & R@1 & R@5 & R@10 & R@1 & R@5 & R@10 & R@1 & R@5 & R@10 & R@1 & R@5 & R@10 \\
    \Xhline{1pt}
    \multirow{6}{*}{CLIP~\cite{radford2021learning}} & CC3M\cite{sharma2018conceptual}  & 26.0 & 52.3 & 64.2 & 18.3 & 42.8 & 55.4 & 40.0 & 67.0 & 76.9 & 28.4 & 56.0 & 66.6  \\
                  & \multirow{2}{*}{Ours}   &   \bf47.3 & \bf  74.1  & \bf  83.2  & \bf  33.2  & \bf62.0    & \bf  73.5  & \bf  75.0  & \bf   92.3  &  \bf95.9  & \bf  58.0  & \bf  83.0  & \bf  89.2   \\
                  & & \footnotesize\textcolor{mygreen}{+21.3}& \footnotesize\textcolor{mygreen}{+21.8}& \footnotesize\textcolor{mygreen}{+19.0}& \footnotesize\textcolor{mygreen}{+14.9}& \footnotesize\textcolor{mygreen}{+19.2}& \footnotesize\textcolor{mygreen}{+18.1}& \footnotesize\textcolor{mygreen}{+35.0}& \footnotesize\textcolor{mygreen}{+25.3}& \footnotesize\textcolor{mygreen}{+19.0}& \footnotesize\textcolor{mygreen}{+29.6}& \footnotesize\textcolor{mygreen}{+27.0}& \footnotesize\textcolor{mygreen}{+22.6}\\
                  & CC3M~\cite{sharma2018conceptual}  & \cellcolor{blue!10}8.7 &  \cellcolor{blue!10}23.9 & \cellcolor{blue!10}33.7 & \cellcolor{blue!10}7.1 & \cellcolor{blue!10}19.7 & \cellcolor{blue!10}28.6 & \cellcolor{blue!10}17.4 & \cellcolor{blue!10}37.9 &  \cellcolor{blue!10}50.1 & \cellcolor{blue!10}13.9 & \cellcolor{blue!10}30.8 &  \cellcolor{blue!10}40.5  \\
                  & \multirow{2}{*}{Ours}&  \bf  \cellcolor{blue!10}35.9  &  \bf  \cellcolor{blue!10}62.4  & \bf   \cellcolor{blue!10}73.9  &  \bf  \cellcolor{blue!10}26.5  &  \bf  \cellcolor{blue!10}51.1  & \bf   \cellcolor{blue!10}62.7  & \bf   \cellcolor{blue!10}63.5  &  \bf  \cellcolor{blue!10}86.6  &   \cellcolor{blue!10}\bf91.7  &  \bf  \cellcolor{blue!10}49.3  & \bf   \cellcolor{blue!10}74.8  &  \bf  \cellcolor{blue!10}83.1   \\
                  & & \cellcolor{blue!10}\footnotesize\textcolor{mygreen}{+27.2}& \cellcolor{blue!10}\footnotesize\textcolor{mygreen}{+38.5}& \cellcolor{blue!10}\footnotesize\textcolor{mygreen}{+40.2}& \cellcolor{blue!10}\footnotesize\textcolor{mygreen}{+19.4}& \cellcolor{blue!10}\footnotesize\textcolor{mygreen}{+31.4}& \cellcolor{blue!10}\footnotesize\textcolor{mygreen}{+34.1}& \cellcolor{blue!10}\footnotesize\textcolor{mygreen}{+46.1}& \cellcolor{blue!10}\footnotesize\textcolor{mygreen}{+48.7}& \cellcolor{blue!10}\footnotesize\textcolor{mygreen}{+41.6}& \cellcolor{blue!10}\footnotesize\textcolor{mygreen}{+35.4}& \cellcolor{blue!10}\footnotesize\textcolor{mygreen}{+44.0}& \cellcolor{blue!10}\footnotesize\textcolor{mygreen}{+42.6}\\
                  \hline
    \multirow{6}{*}{BLIP~\cite{li2022blip}} & CC3M\cite{sharma2018conceptual}  & 67.5 & 88.9 & 93.7 & 50.9 & 76.9 & 85.2 & 88.5 & 98.5 &  99.5 & 74.0 & 92.4 & 95.7  \\
                  & \multirow{2}{*}{Ours}   &   \bf76.2 & \bf 93.0 & \bf 96.4 & \bf 57.6 & \bf 82.3 & \bf 89.3 & \bf 94.1 & \bf  99.4 & \bf99.8 & \bf 82.2 & \bf 95.7 & \bf 97.9  \\
                  & & \footnotesize\textcolor{mygreen}{+8.7}& \footnotesize\textcolor{mygreen}{+4.1}& \footnotesize\textcolor{mygreen}{+2.7}& \footnotesize\textcolor{mygreen}{+6.7}& \footnotesize\textcolor{mygreen}{+5.4}& \footnotesize\textcolor{mygreen}{+4.1}& \footnotesize\textcolor{mygreen}{+5.6}& \footnotesize\textcolor{mygreen}{+0.9}& \footnotesize\textcolor{mygreen}{+0.3}& \footnotesize\textcolor{mygreen}{+8.2}& \footnotesize\textcolor{mygreen}{+3.3}& \footnotesize\textcolor{mygreen}{+2.2}\\
                  & CC3M~\cite{sharma2018conceptual}  & \cellcolor{blue!10}36.3 &\cellcolor{blue!10}  62.6 &\cellcolor{blue!10} 73.6 &\cellcolor{blue!10} 28.6 &\cellcolor{blue!10} 52.9 &\cellcolor{blue!10} 64.2 &\cellcolor{blue!10} 62.1 &\cellcolor{blue!10} 87.4 &\cellcolor{blue!10} 91.9 &\cellcolor{blue!10} 51.2 &\cellcolor{blue!10} 75.5 &\cellcolor{blue!10}  82.7  \\
                  & \multirow{2}{*}{Ours}&\cellcolor{blue!10}  \bf \text{59.3} &\cellcolor{blue!10}  \bf 81.5 &\cellcolor{blue!10} \bf  88.9 &\cellcolor{blue!10}  \bf 45.2 &\cellcolor{blue!10}  \bf 71.0 &\cellcolor{blue!10} \bf  80.2 &\cellcolor{blue!10} \bf  85.5 &\cellcolor{blue!10}  \bf 97.1 &\cellcolor{blue!10}  \bf98.8 &\cellcolor{blue!10}  \bf 72.0 &\cellcolor{blue!10} \bf  90.3 &\cellcolor{blue!10}  \bf 94.3  \\
                  & &\cellcolor{blue!10} \footnotesize\textcolor{mygreen}{+23.0}&\cellcolor{blue!10} \footnotesize\textcolor{mygreen}{+18.9}&\cellcolor{blue!10} \footnotesize\textcolor{mygreen}{+15.3}& \cellcolor{blue!10}\footnotesize\textcolor{mygreen}{+16.8}& \cellcolor{blue!10}\footnotesize\textcolor{mygreen}{+18.1}& \cellcolor{blue!10}\footnotesize\textcolor{mygreen}{+16.0}& \cellcolor{blue!10}\footnotesize\textcolor{mygreen}{+23.4}& \cellcolor{blue!10}\footnotesize\textcolor{mygreen}{+9.7}& \cellcolor{blue!10}\footnotesize\textcolor{mygreen}{+6.9}& \cellcolor{blue!10}\footnotesize\textcolor{mygreen}{+20.8}& \cellcolor{blue!10}\footnotesize\textcolor{mygreen}{+14.8}& \cellcolor{blue!10}\footnotesize\textcolor{mygreen}{+11.6}\\\hline
    \end{tabular}}
    
    \label{tab:image-text-retrieval}
\end{table*}

%% file: tables/clip_zeroshot.tex
\begin{table*}[t]
\small
\centering
\caption{The zero-shot evaluation results of our method on 15 common classification datasets and ImageNet with CLIP pre-trained on CC3M and CC12M. The best results are highlighted in \textbf{bold}. On CC3M, our method demonstrates an average improvement of 13.4 on 15 datasets and 13.1 on ImageNet. The different results on vanilla CLIP are due to the different training data downloaded from the Internet.}
\vspace*{1.5mm}
\label{table:zeroshot-main}
\resizebox{1.0\textwidth}{!}{
\begin{tabular}
{c@{\hspace{1.1em}}c@{\hspace{0.5em}}|c@{\hspace{0.5em}}c@{\hspace{0.5em}}c@{\hspace{0.5em}}c@{\hspace{0.5em}}c@{\hspace{0.5em}}c@{\hspace{0.5em}}c@{\hspace{0.5em}}c@{\hspace{0.5em}}c@{\hspace{0.5em}}c@{\hspace{0.5em}}c@{\hspace{0.5em}}c@{\hspace{0.5em}}c@{\hspace{0.5em}}c@{\hspace{0.5em}}c@{\hspace{0.5em}}c@{\hspace{0.5em}}c@{\hspace{0.5em}}|c@{\hspace{0.5em}}c@{\hspace{0.5em}}}

\toprule[1.2pt]
\bf Data&\bf Model&
\rotatebox[origin=lb]{90}{\smash{Food-101}} & \rotatebox[origin=lb]{90}{\smash{CIFAR-10}} & \rotatebox[origin=lb]{90}{\smash{CIFAR-100}} & \rotatebox[origin=lb]{90}{\smash{SUN397}} &
\rotatebox[origin=lb]{90}{\smash{Cars}} & \rotatebox[origin=lb]{90}{\smash{Aircraft}} & \rotatebox[origin=lb]{90}{\smash{DTD}} & \rotatebox[origin=lb]{90}{\smash{Pets}} & \rotatebox[origin=lb]{90}{\smash{Caltech-101}} &
\rotatebox[origin=lb]{90}{\smash{Flowers}} & \rotatebox[origin=lb]{90}{\smash{STL-10}} & \rotatebox[origin=lb]{90}{\smash{EuroSAT}} &
\rotatebox[origin=lb]{90}{\smash{RESISC45}} & \rotatebox[origin=lb]{90}{\smash{GTSRB}} & \rotatebox[origin=lb]{90}{\smash{Country211}}  & \rotatebox[origin=lb]{90}{\smash{\colorbox[HTML]{fee3c8}{\bf Average}}} & \rotatebox[origin=lb]{90}{\smash{\colorbox[HTML]{c0ecfa}{\bf $\Delta$(\%)}}} & \rotatebox[origin=lb]{90}{\smash{\colorbox[HTML]{fee3c8}{\bf ImageNet}}} & \rotatebox[origin=lb]{90}{\smash{\colorbox[HTML]{c0ecfa}{\bf $\Delta$(\%)}}}\\
\midrule
\multicolumn{19}{c}{\textit{Model Architecture: ViT-B/16}}\\
\midrule
\multirow{4}{2.5em}{\rotatebox[origin=c]{0}{CC3M}}  & CLIP & 10.3 & 54.9 & 21.8 & 25.0 & 0.8  & 1.4 & 10.5 & 12.8 & 43.3 & 10.2 & 77.6 & 14.1 & 19.1 &  6.9  & 0.6 & \colorbox[HTML]{fee3c8}{20.6} & - &\colorbox[HTML]{fee3c8}{15.8} & -\\
                                                    &  LaCLIP~\cite{fan2023improving} &  14.2 &  57.1 &  27.5 &  35.1 &  1.6  & \bf 1.6 &  16.6 &  15.6 &  52.7 &  14.7 &  86.2 &  15.0 & \bf 24.3 & 6.4  &  1.0 &  \colorbox[HTML]{fee3c8}{24.6} & \colorbox[HTML]{c0ecfa}{$\uparrow$4.0} & \colorbox[HTML]{fee3c8}{21.5} & \colorbox[HTML]{c0ecfa}{$\uparrow$5.7}\\
                                                    \cmidrule{2-21}
                                                    &  CLIP & 9.2& 28.4 & 9.8 & 21.9 & 1.1 & 1.1 & 8.3 & 7.6 & 40.7 & 8.9 & 70.4 & 13.2&14.8 & 5.6 & 0.4 & \colorbox[HTML]{fee3c8}{16.1} & - & \colorbox[HTML]{fee3c8}{11.9} & -\\
                                                    & Ours & \bf18.7  & \bf58.4  & \bf32.4  & \bf43.8  & \bf3.9  & 1.5  & \bf20.2  & \bf32.1  & \bf63.5  & \bf17.5  & \bf87.3  & \bf25.1 & 23.1  & \bf13.0  & \bf2.0  & \colorbox[HTML]{fee3c8}{\bf29.5}  & \colorbox[HTML]{c0ecfa}{$\uparrow$ \bf13.4} &\colorbox[HTML]{fee3c8}{\bf25.0}   & \colorbox[HTML]{c0ecfa}{$\uparrow$ \bf13.1}\\
\midrule
\multirow{4}{2.8em}{\rotatebox[origin=c]{0}{CC12M}}  & CLIP & 50.8 & 64.9 & 38.5 & 44.7 & 24.1 & 2.4 & 19.4 & 64.1 & 77.4 & 33.2 & 91.0 & 20.1 & 38.9 & 7.3  & 5.1 & \colorbox[HTML]{fee3c8}{38.8} & - &\colorbox[HTML]{fee3c8}{40.2} & -\\
                                                    & LaCLIP~\cite{fan2023improving} &  60.7 &  75.1 &  43.9 &  57.0 & \bf 36.3 & \bf 5.6 & \bf 31.0 & \bf 72.4 & \bf 83.3 & \bf 39.9 &  95.1 &  27.3 &  44.3 &  12.7 & \bf 8.9 & \colorbox[HTML]{fee3c8}{ 46.2} & \colorbox[HTML]{c0ecfa}{$\uparrow$7.4} &\colorbox[HTML]{fee3c8}{\bf 48.4} & \colorbox[HTML]{c0ecfa}{$\uparrow$8.2}\\
                                                    \cmidrule{2-21}
                                                    &  CLIP & 45.9& 65.7 & 33.4 & 44.7 & 18.4 & 2.9 & 18.5 & 54.8 & 72.6 & 30.4 & 89.5 & 23.2  & 28.5 & 9.3 & 4.8 & \colorbox[HTML]{fee3c8}{36.2} & - &\colorbox[HTML]{fee3c8}{37.3} & -\\
                                                    &  Ours & \bf60.9  & \bf83.0  & \bf55.4  & \bf59.4  & 24.1  & 3.2  & 30.7  & 64.8  & 79.3  & 36.0  & \bf95.3  & \bf40.5 & \bf45.6  & \bf25.4  & 5.8  & \colorbox[HTML]{fee3c8}{\bf 47.3} &  \colorbox[HTML]{c0ecfa}{$\uparrow$ \bf11.1} &\colorbox[HTML]{fee3c8}{47.5}  & \colorbox[HTML]{c0ecfa}{$\uparrow$ \bf10.2}  \\
\bottomrule[1.2pt]
\end{tabular}
}
\end{table*}

%% file: tables/linear_probing.tex
\begin{table*}[t]
\small
\centering
\vspace*{1.0cm}
\caption{The linear probing results of our method on 15 common classification datasets with CLIP pre-trained on CC3M and CC12M. On CC3M, our method demonstrates an average improvement of 7.9 on 15 datasets.}
\vspace*{1.5mm}
\label{table:linear-main}
\resizebox{1.0\textwidth}{!}{
\begin{tabular}
{c@{\hspace{1.1em}}c@{\hspace{0.5em}}|c@{\hspace{0.5em}}c@{\hspace{0.5em}}c@{\hspace{0.5em}}c@{\hspace{0.5em}}c@{\hspace{0.5em}}c@{\hspace{0.5em}}c@{\hspace{0.5em}}c@{\hspace{0.5em}}c@{\hspace{0.5em}}c@{\hspace{0.5em}}c@{\hspace{0.5em}}c@{\hspace{0.5em}}c@{\hspace{0.5em}}c@{\hspace{0.5em}}c@{\hspace{0.5em}}c@{\hspace{0.5em}}|c@{\hspace{0.5em}}}

\toprule[1.2pt]
\bf Data&\bf Model&
\rotatebox[origin=lb]{90}{\smash{Food-101}} & \rotatebox[origin=lb]{90}{\smash{CIFAR-10}} & \rotatebox[origin=lb]{90}{\smash{CIFAR-100}} & \rotatebox[origin=lb]{90}{\smash{SUN397}} &
\rotatebox[origin=lb]{90}{\smash{Cars}} & \rotatebox[origin=lb]{90}{\smash{Aircraft}} & \rotatebox[origin=lb]{90}{\smash{DTD}} & \rotatebox[origin=lb]{90}{\smash{Pets}} & \rotatebox[origin=lb]{90}{\smash{Caltech-101}} &
\rotatebox[origin=lb]{90}{\smash{Flowers}} & \rotatebox[origin=lb]{90}{\smash{STL-10}} & \rotatebox[origin=lb]{90}{\smash{EuroSAT}} &
\rotatebox[origin=lb]{90}{\smash{RESISC45}} & \rotatebox[origin=lb]{90}{\smash{GTSRB}} & \rotatebox[origin=lb]{90}{\smash{Country211}}  & \rotatebox[origin=lb]{90}{\smash{\colorbox[HTML]{fee3c8}{\bf Average}}} & \rotatebox[origin=lb]{90}{\smash{\colorbox[HTML]{c0ecfa}{\bf $\Delta$(\%)}}}\\
\midrule
\multicolumn{19}{c}{\textit{Model Architecture: ViT-B/16}}\\
\midrule
\multirow{4}{2.5em}{\rotatebox[origin=c]{0}{CC3M}}  & CLIP & 62.6 & 86.8 & 68.1 & 58.5 & 32.8  & 40.9 & 63.4 & 69.6 & 82.0 & 89.4 & 91.7 & 95.9 & 89.0 &  71.9  & 13.3 & \colorbox[HTML]{fee3c8}{67.7} & -\\
                                                    &  LaCLIP~\cite{fan2023improving} &  63.8 &  87.7 &  69.5 &  60.2 &  32.4  &  42.7 &  64.0 &  71.1 &  83.3 &  90.2 &  93.4 &  95.8 &  89.7 & 74.6  &  13.2 &  \colorbox[HTML]{fee3c8}{68.8} & \colorbox[HTML]{c0ecfa}{$\uparrow$1.1}\\
                                                    \cmidrule{2-19}
                                                    &  CLIP & 52.8& 79.7 & 58.8 & 52.4 & 21.9 & 26.4 & 53.1 & 56.6 & 77.5 & 76.6 & 85.7 & 93.9&84.5 & 66.7 & 8.7 & \colorbox[HTML]{fee3c8}{59.7} & -\\
                                                    & Ours & 64.0  & 87.7  & 68.5  & 59.1  & 34.5  & 32.1  & 60.4  & 73.3  & 85.5  & 83.6  & 92.6  & 95.3 & 87.9  & 78.4  & 10.6  & \colorbox[HTML]{fee3c8}{67.6}  & \colorbox[HTML]{c0ecfa}{$\uparrow$ \bf7.9} \\
\midrule
\multirow{4}{2.8em}{\rotatebox[origin=c]{0}{CC12M}}  & CLIP & 81.6 & 93.8 & 79.3 & 72.0 & 75.1 & 52.6 & 75.6 & 86.2 & 92.2 & 95.3 & 97.3 & 96.7 & 93.1 & 80.6  & 19.7 & \colorbox[HTML]{fee3c8}{79.4} & - \\
                                                    & LaCLIP~\cite{fan2023improving} &  82.9 &  94.7 &  79.7&  73.8 & 79.9 &  54.5 &  75.7 &  87.7 &  93.0 &  96.4 &  98.0 &  96.4 &  93.0 &  81.9 & 19.7 & \colorbox[HTML]{fee3c8}{80.5} & \colorbox[HTML]{c0ecfa}{$\uparrow$1.1} \\
                                                    \cmidrule{2-19}
                                                    &  CLIP & 73.9& 89.9 & 71.2 & 68.9 & 62.8 & 35.6 & 67.1 & 81.1 & 90.3 & 87.7 & 95.0 & 95.6  & 89.1 & 77.4 & 13.8 & \colorbox[HTML]{fee3c8}{73.3} & - \\
                                                    &  Ours & 80.6  & 93.9  & 78.1  & 72.1  & 62.1  & 35.6  & 72.1  & 82.7  & 90.6  & 87.9  & 97.3  & 95.7 & 91.1  & 84.8  & 13.8  & \colorbox[HTML]{fee3c8}{ 75.9} &  \colorbox[HTML]{c0ecfa}{$\uparrow$ \bf2.6} \\
\bottomrule[1.2pt]
\end{tabular}
}
\end{table*}

%% file: tables/cc3m_othertasks.tex
\begin{table*}[t]
	\centering	
        \footnotesize
        \caption
	{
        The results of vision question answering, visual reasoning and image captioning of BLIP pre-trained on CC3M, CC12M, and YFCC15M, respectively.
	}
    \resizebox{1.0\textwidth}{!}{
	\begin{tabular}	{l|llllllllll}
	 \multirow[t]{2}{*}{Dataset}&  \multicolumn{2}{c}{VQAv2}  &A-OKVQA &OK-VQA & \multicolumn{2}{c}{NLVR$^2$} &\multicolumn{2}{c}{COCO Caption} &\multicolumn{2}{c}{NoCaps}\\
	  & test-dev & test-std &   val & test &dev & test-P & B@4 & CIDEr &CIDEr&SPICE\\
	  		\shline 	
	   CC3M~\cite{sharma2018conceptual}& 71.5 & 71.8  &28.9 & 24.0& 76.0 & 76.2&36.6&119.9&84.4&12.5 \\
    Ours & 75.6$_{\textcolor{mygreen}{+4.1}}$ & 75.6$_{\textcolor{mygreen}{+3.8}}$ & 32.0$_{\textcolor{mygreen}{+3.1}}$& 31.4$_{\textcolor{mygreen}{+7.4}}$ & 80.1$_{\textcolor{mygreen}{+4.1}}$ &79.3$_{\textcolor{mygreen}{+3.1}}$&37.6$_{\textcolor{mygreen}{+1.0}}$&125.6$_{\textcolor{mygreen}{+5.7}}$&97.0$_{\textcolor{mygreen}{+12.6}}$&13.8$_{\textcolor{mygreen}{+1.3}}$\\
    \midrule
    CC12M~\cite{changpinyo2021conceptual}& 73.5 & 73.5  & 31.5 & 31.0 &78.7 & 79.0& 37.2& 123.1& 99.4&13.5 \\
    Ours & 77.0$_{\textcolor{mygreen}{+3.5}}$ & 77.1$_{\textcolor{mygreen}{+3.6}}$  & 32.0$_{\textcolor{mygreen}{+0.5}}$ & 31.6$_{\textcolor{mygreen}{+0.6}}$ & 81.2$_{\textcolor{mygreen}{+2.5}}$ &81.9$_{\textcolor{mygreen}{+2.9}}$&39.1$_{\textcolor{mygreen}{+1.9}}$&130.1$_{\textcolor{mygreen}{+7.0}}$&103.6$_{\textcolor{mygreen}{+4.2}}$&14.3$_{\textcolor{mygreen}{+0.8}}$\\
    \midrule
    YFCC15M~\cite{thomee2016yfcc100m}& 70.5 & 70.6  & 29.8 & 29.2 &74.0 & 74.2& 36.8& 122.0& 97.5& 13.3\\
    Ours & 72.8$_{\textcolor{mygreen}{+2.3}}$ & 72.8$_{\textcolor{mygreen}{+2.2}}$ &31.1$_{\textcolor{mygreen}{+1.3}}$& 30.0$_{\textcolor{mygreen}{+0.8}}$& 78.5$_{\textcolor{mygreen}{+4.5}}$ &77.8$_{\textcolor{mygreen}{+3.6}}$&37.6$_{\textcolor{mygreen}{+0.8}}$&125.6$_{\textcolor{mygreen}{+3.6}}$&101.1$_{\textcolor{mygreen}{+3.6}}$&13.9$_{\textcolor{mygreen}{+0.6}}$

	\end{tabular}}
    \vspace{-1em}
     	
	\label{tab:vqa_nlvr}
\end{table*}		

%% file: tables/compared_veclip.tex
\definecolor{ForestGreen}{rgb}{0.13, 0.55, 0.13}
\definecolor{Green}{rgb}{0.0, 0.5, 0.0}
\definecolor{green(munsell)}{rgb}{0.0, 0.66, 0.47}
\definecolor{green(ryb)}{rgb}{0.4, 0.69, 0.2}
\definecolor{green(pigment)}{rgb}{0.0, 0.65, 0.31}

\newcommand{\cmark}{\text{\ding{51}}}%
\newcommand{\xmark}{\text{\ding{55}}}%

\newcommand{\tablestyle}[2]{\setlength{\tabcolsep}{#1}\renewcommand{\arraystretch}{#2}\centering}
\newcommand{\km}[1]{{\color{red}[km: #1]}}
\newcommand{\rbg}[1]{{\color{blue}[rbg: #1]}}
\newcommand{\ppd}[1]{{\color{green}[ppd: #1]}}
\newcommand{\bd}[1]{\textbf{#1}}
\newcommand{\app}{\raise.17ex\hbox{$\scriptstyle\sim$}}
\newcommand{\ncdot}{{\mkern 0mu\cdot\mkern 0mu}}

\newcommand{\green}[1]{\multicolumn{1}{c}{\color{green(pigment)}#1}}
\newcommand{\red}[1]{\multicolumn{1}{c}{\color{red}#1}}
\newcommand{\cell}[1]{\multicolumn{1}{r}{#1}}

\begin{table*}[ht]
\centering
\caption{
\small The image-text retrieval results on COCO and Flickr30K and classification results on ImageNet and ImageNetV2~\cite{recht2019imagenet} using the pre-trained CLIP on CC3M.}
\label{table:VeCLIP}
{
\centering
\resizebox{0.8\textwidth}{!}{
\begin{tabular}{c|ccccccc}
\toprule[1.2pt]
\multirow{2}{*}{\bf Method} & \multicolumn{2}{c}{ \bf \quad COCO (R@1) \quad} & \multicolumn{2}{c}{\bf Flickr30k (R@1)}  & \multirow{2}{*}{ \bf ImageNet}  & \multirow{2}{*}{ \bf ImageNetV2} & \multirow{2}{*}{ \bf Avg.} \\
 & I2T  & T2I & I2T  & T2I   &      &   \\

\hline
CLIP  &  13.9 & 9.6  & 26.3  & 18.0  & 14.6  &  12.5    &  15.8\\
VeCLIP~\cite{lai2023scarcity}   & 32.0 & 22.1 & 57.2 &  36.5 & 20.7 & 17.9    & 31.1\\
\hline
CLIP  &  8.7 & 7.1  & 17.4  & 13.9  & 11.9  &  10.3    &  11.6\\
Ours   & \bf 35.9 & \bf 26.5 & \bf 63.5 & \bf 49.3 & \bf 25.0 & \bf 21.4   & \bf36.9\\
\bottomrule[1.2pt]
\end{tabular}}
}

\end{table*}

%% file: sec/5_conclusion.tex
\section{Conclusion and Discussion}
In this paper, we propose to augment visual-language representation learning by leveraging multiple MLLMs. While retaining the rich visual information of the original dataset, we utilize multiple MLLMs to extend diverse captions for each image. Additionally, we introduce ``text shearing" to address issues of hallucinations and monotonous language style in synthetic captions. Validated across various visual-language pre-training frameworks and datasets, our method significantly improves performance on numerous downstream tasks. 
This encourages further exploration in the utilization of MLLMs in the future.

\textbf{Limitations and future works}
Although our method has achieved remarkable results in enhancing visual-language representation learning, a certain proportion of noise persists due to unreliable MLLMs' outputs.
These noises limit the further improvement of the model's performance.
Future research could explore leveraging more powerful MLLMs to generate accurate captions and expanding to larger datasets.

%% file: sec/supp.tex
\section{Dataset Details}
\textbf{Evaluation datasets.} We provide specific information on the datasets used in the image classification task in Table~\ref{tab:classification} and the datasets used in the multi-modal task in Table~\ref{tab:multimodal_datasets}.
\input{tables/classification_datasets}
\input{tables/multimodal_datasets}
\section{Implementation Details}
\subsection{Caption Generation}
We provide the hyper-parameter settings of the used MLLMs: \{MiniGPT-4~\cite{zhu2023minigpt}, Otter~\cite{li2023otter}, Qwen-VL~\cite{bai2023qwen}, LLaVA-1.5~\cite{liu2023improved}\} in the Table~\ref{tab:parameter}. After the new tokens are generated and decoded using the decoder, we extract the first meaningful substring in the string whose length is greater than 5 and ends with a period. Then this substring is extended into the caption list for each image.
\input{tables/parameter}

\subsection{Evaluation}
When evaluating image-text retrieval tasks, our settings are as follows: In the zero-shot retrieval of MSCOCO and Flickr30K, we directly use pre-trained models. Unlike the zero-shot retrieval of Flickr30K in BLIP~\cite{li2022blip}, we do not use a model that is pre-trained and fine-tuned on MSCOCO. Correspondingly, the results of fine-tuning retrieval are the results of retrieval after fine-tuning on the two target datasets respectively.

\section{Comparison with More Methods}
We also compare our work with the method mentioned in the~\cite{nguyen2023improving}. 
Regarding the methodology, ~\cite{nguyen2023improving} employs a similar approach of combining raw captions and generated captions for training. In contrast to our use of multiple advanced MLLMs, ~\cite{nguyen2023improving} relies solely on BLIP2~\cite{li2023blip} for rewriting captions. While the aim of ~\cite{nguyen2023improving} is to eliminate noise in the original large-scale dataset, our method aims to enhance visual-language representation learning. Concerning experimental settings, we primarily conduct pre-training based on the two datasets CC3M and CC12M, whereas ~\cite{nguyen2023improving} focuses on pre-training using DataComp~\cite{gadre2023datacomp} datasets of 12.8M, 128M, and 1.28B. Regarding experimental results, our pre-trained CLIP on CC12M achieved a zero-shot classification result of 47.5 on ImageNet, surpassing even the best result of 31.7 on the medium scale (128M). This further affirms the effectiveness of our method.

\section{Statistics of Words in MLLMs} 
\begin{figure}
    \centering
    \includegraphics[width=0.75\textwidth]{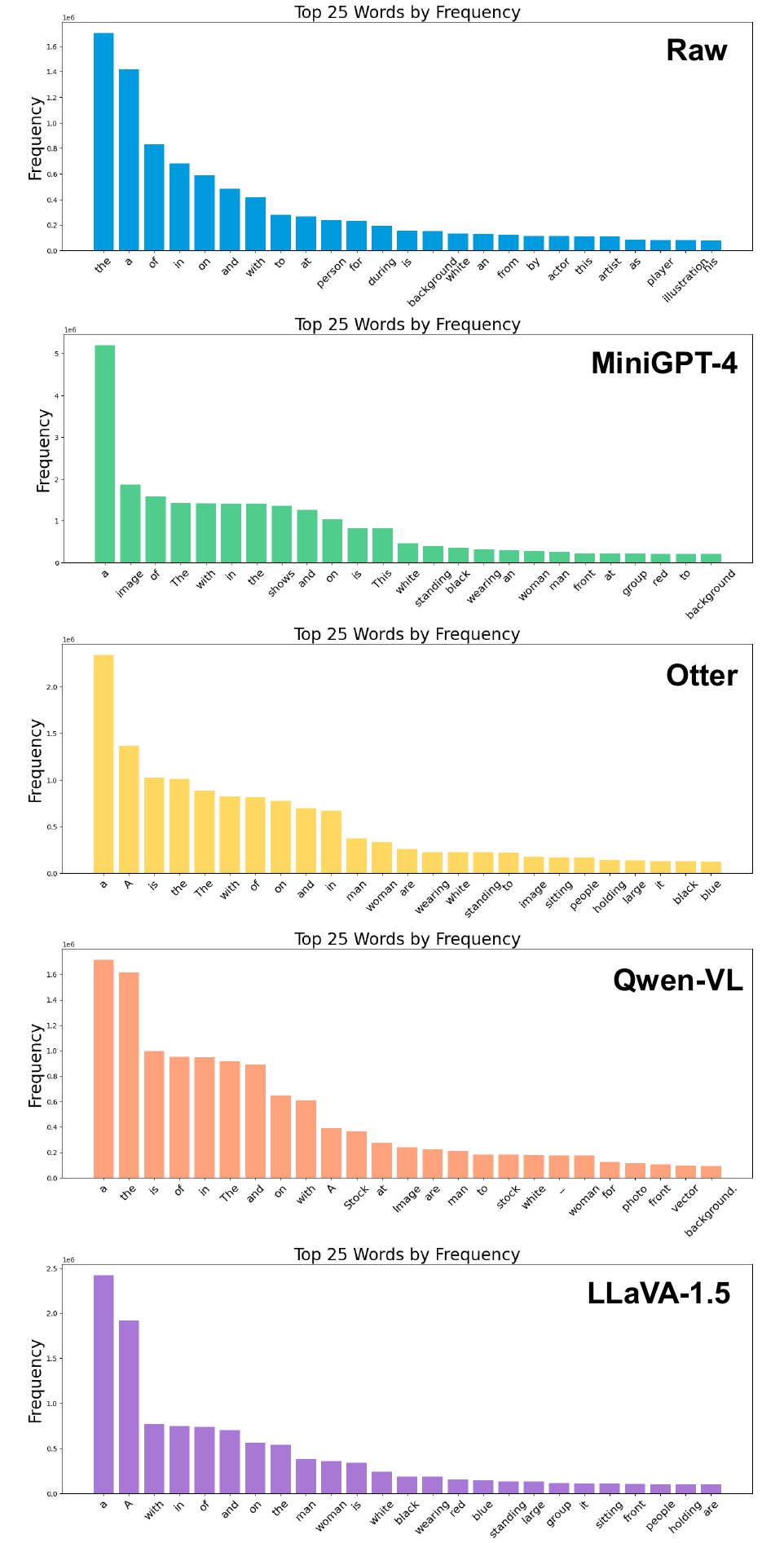}
    \caption{Statistics of commonly used words in MLLMs.}
    \label{fig:commonly_words}
\end{figure}

We count the most commonly used words in captions generated using different MLLMs in Figure~\ref{fig:commonly_words}. The different distributions of these MLLMs on common words also imply the diversity of generated captions.

\section{Details about Code}
\begin{figure*}
    \centering
    \includegraphics[width=0.7\textwidth]{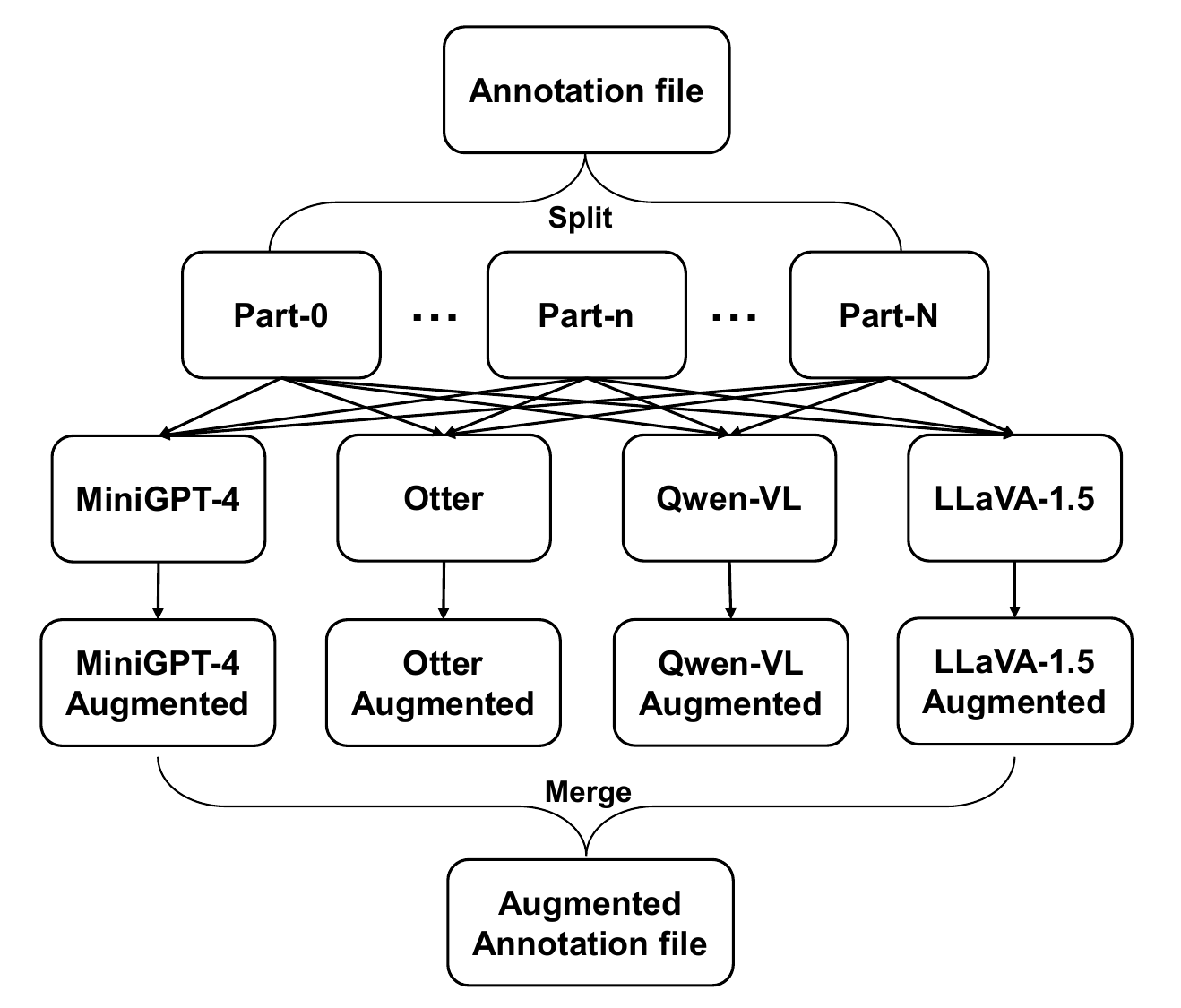}
    \caption{Code running logic.}
    \label{fig:code}
\end{figure*}

We provide our code running logic in Figure~\ref{fig:code}.
The code mainly includes the generation of multi-view captions. For the image-text dataset, we use annotation JSON files and corresponding image data for enhancement. Annotation files include several entries consisting of image paths and captions.
We split the annotation file into multiple parts to improve GPU utilization. For each part, we use MLLMs to generate captions. The captions generated by multiple MLLMs and the original captions are merged again as a new enhanced dataset. Each image in the new dataset contains an original caption and 4 generated captions.

\section{Word Cloud Visualization.}
\begin{figure*}
    \centering
    \includegraphics[width=0.7\textwidth]{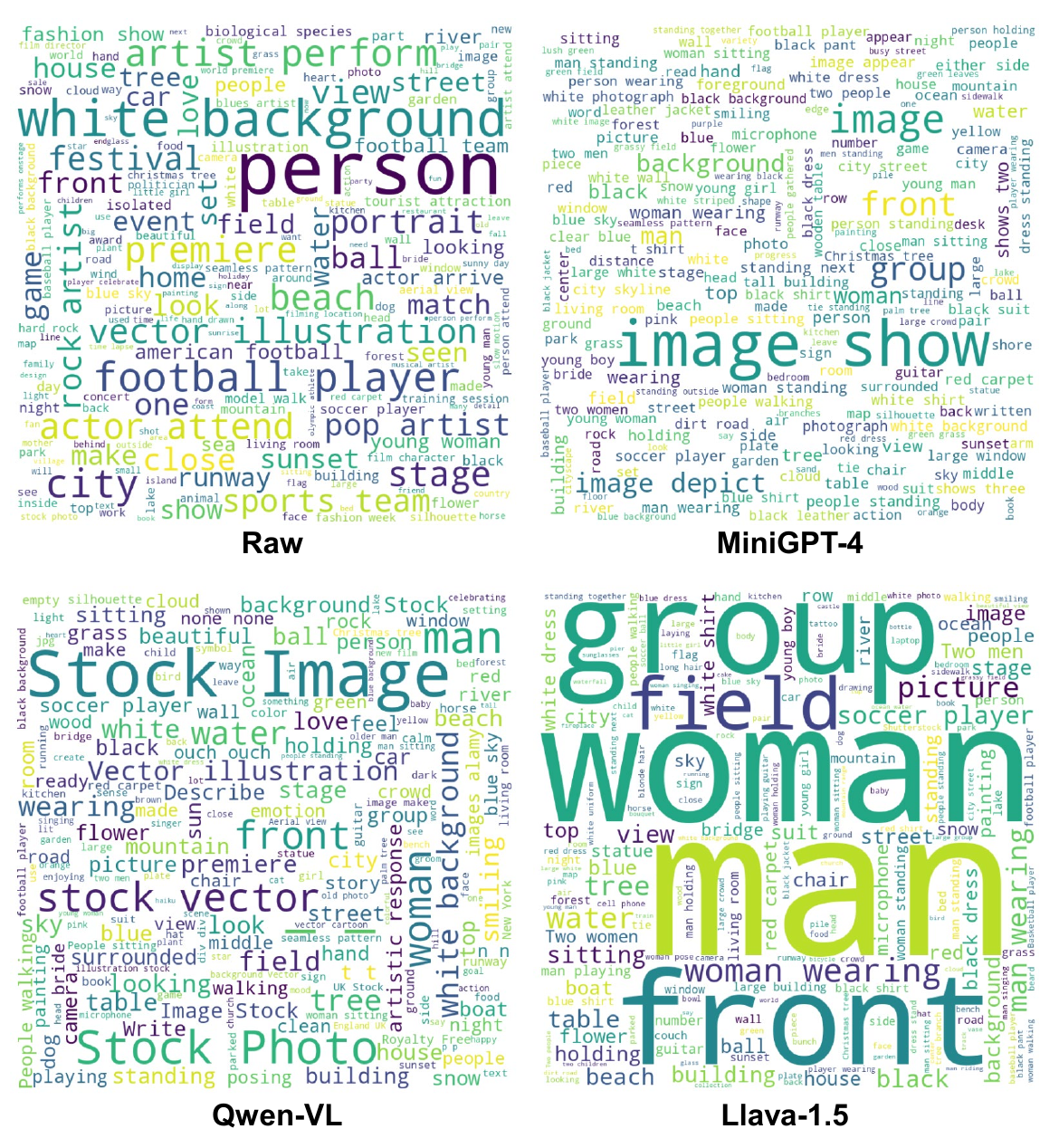}
    \caption{Visualization of word clouds in captions generated by different MLLMs.}
    \label{fig:cloud}
\end{figure*}

We present the word cloud visualization depicting captions generated by various MLLMs in Figure~\ref{fig:cloud}. Notably, there are significant differences in common words.

\section{Image-text Pairs and Image Captioning}
\begin{figure*}
\centering
    \includegraphics[width=\textwidth]{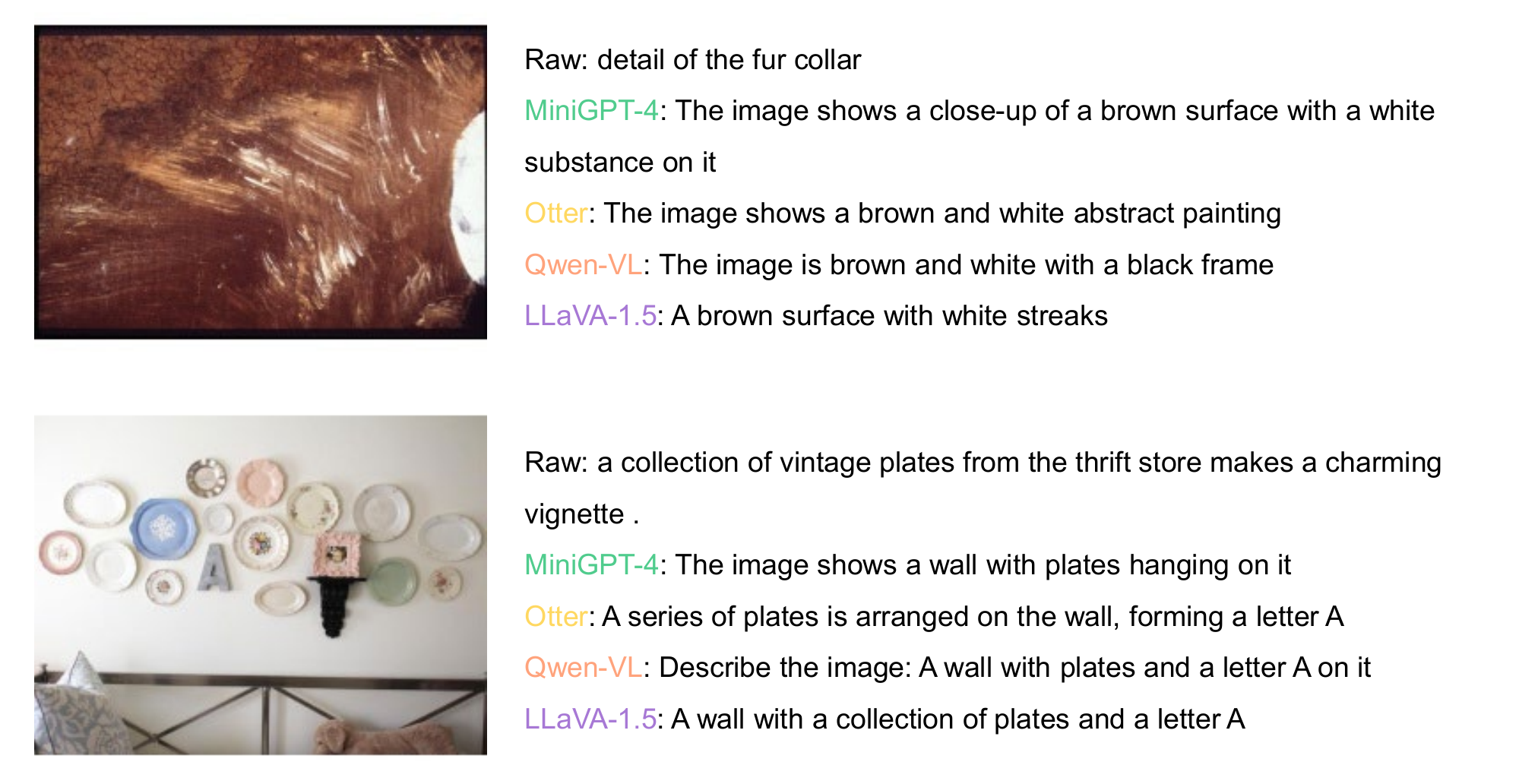}
    \includegraphics[width=\textwidth]{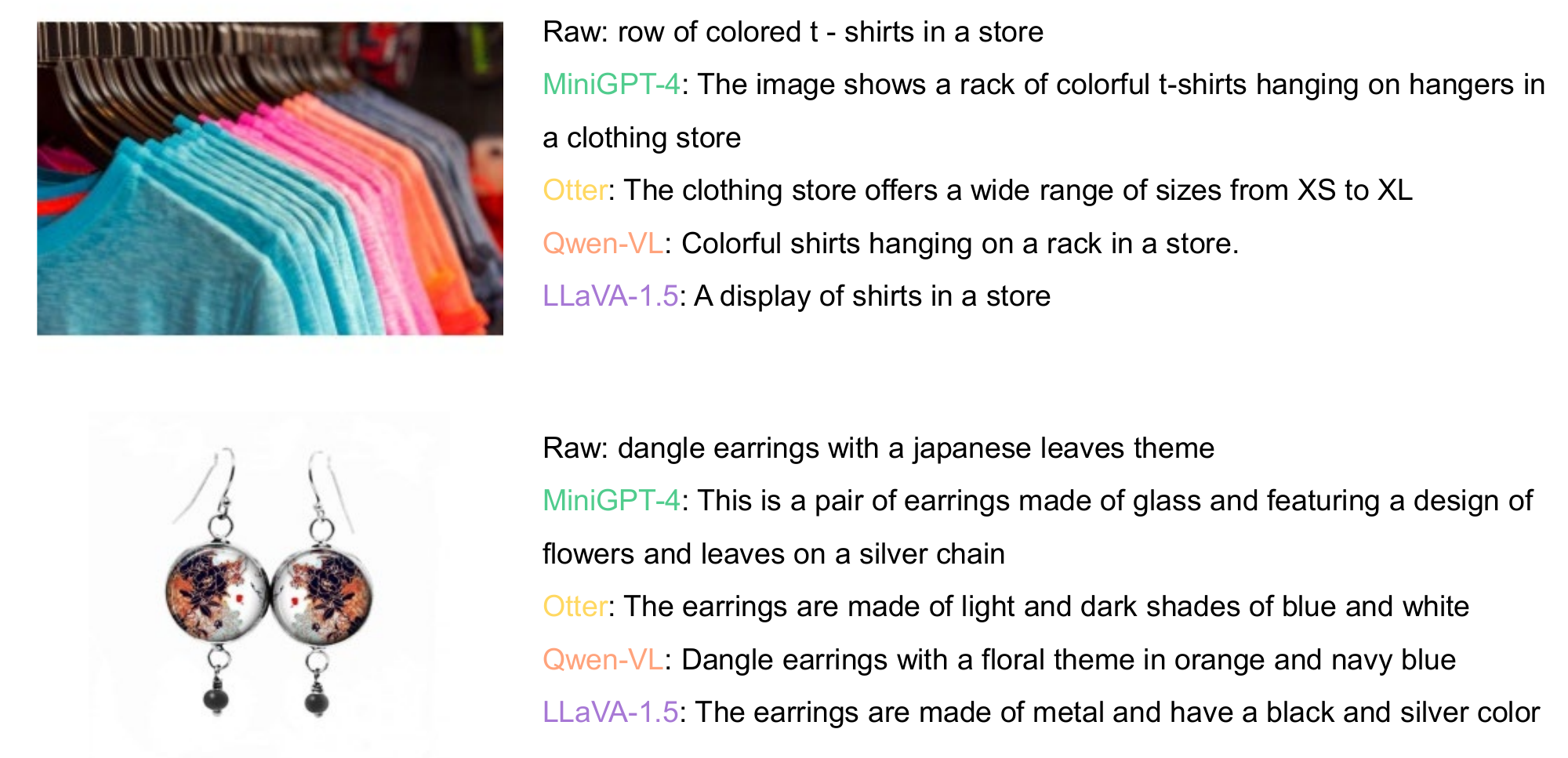}
    \includegraphics[width=\textwidth]{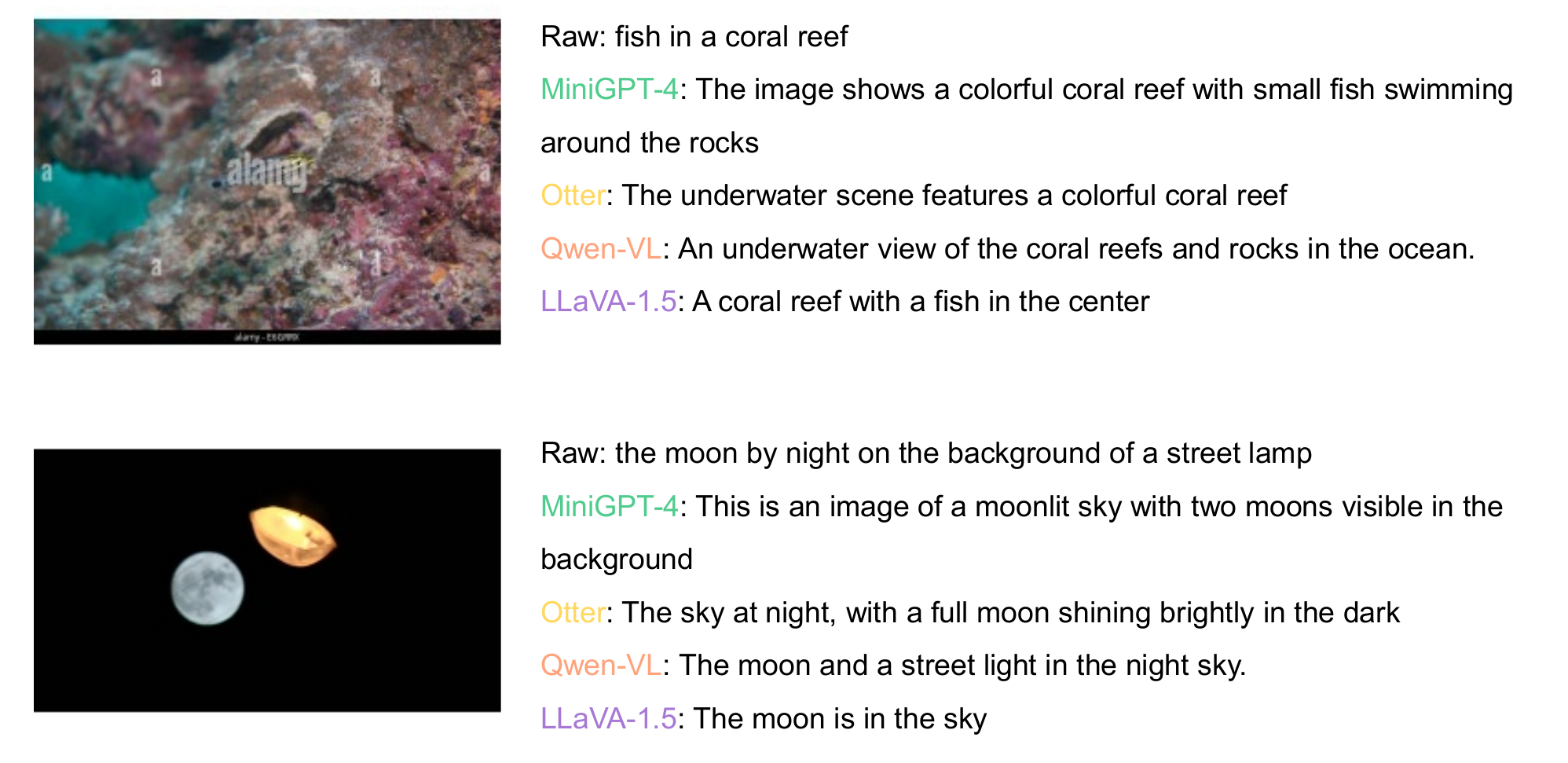}
\caption{Enhanced Image-text pairs. Interestingly, these MLLMs have different image recognition capabilities. The visual concepts extracted by different MLLMs enrich the diversity of captions.}
\label{fig:pairs1}
\end{figure*}

\begin{figure*}
    \includegraphics[width=\textwidth]{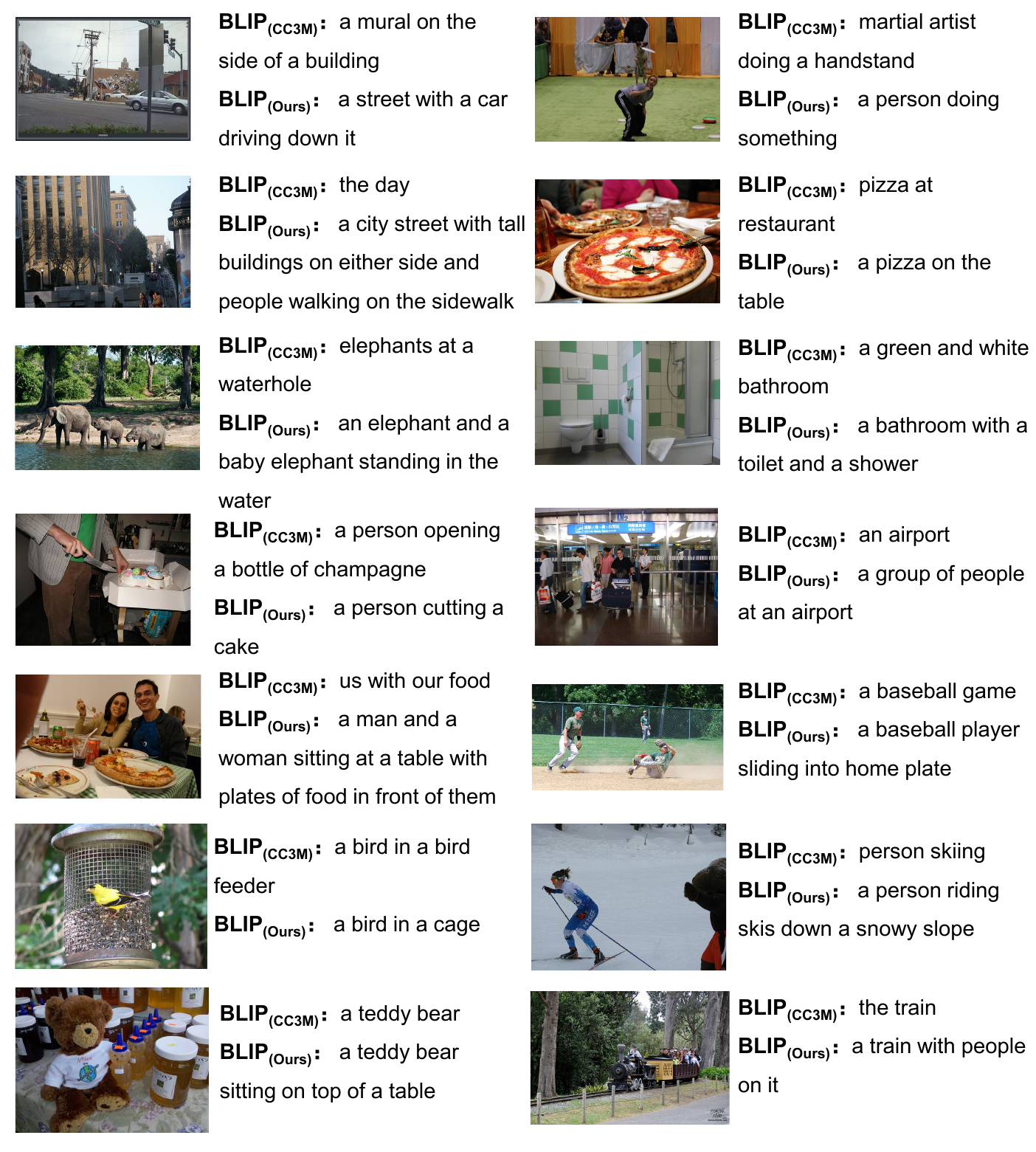}
    \caption{Using pre-trained BLIP to caption the image. Images are from MSCOCO.}
    \label{fig:captioning}
\end{figure*}

We provide more image-text pairs on CC3M enhanced by our method in the Figure~\ref{fig:pairs1}. 
We also use BLIP pre-trained on CC3M without any fine-tuning for image captioning and provide results in Figure~\ref{fig:captioning}. The images used are from MSCOCO.
\clearpage
\clearpage

\clearpage

%% file: tables/classification_datasets.tex
\begin{table}[h]
  \centering
  \caption{The details of the image classification datasets.}
  \resizebox{0.5\textwidth}{!}{
  \begin{tabular}{c|c|c|c}
    \hline
    Dataset & Categories & Train Size & Test Size\\
    \hline
    Food-101~\cite{bossard2014food} & 101 & 75,750&25,250\\
    CIFAR-10~\cite{krizhevsky2009learning} & 10 & 50,000&10,000\\
    CIFAR-100~\cite{krizhevsky2009learning} & 100 & 50,000&10,000\\
    SUN397~\cite{xiao2010sun} & 397 &19,850 &19,850\\
    Cars~\cite{krause20133d} & 196 & 8,144&8,041\\
    Aircraft~\cite{maji2013fine} & 100 & 6,667& 3,333\\
    DTD~\cite{cimpoi2014describing} & 47 & 3,760& 1,880\\
    Pets~\cite{parkhi2012cats} & 37 & 3,680& 3,669\\
    Caltech-101~\cite{fei2006one} & 102 & 3,060& 6,085\\
    Flowers~\cite{nilsback2008automated} & 102 & 2,040& 6,149\\
    STL-10~\cite{coates2011analysis} & 10 & 1,000& 8,000\\
    EuroSAT~\cite{helber2019eurosat} & 10 & 10,000& 5,000\\
    RESISC45~\cite{cheng2017remote} & 45 & 25,200&6,300\\
    GTSRB~\cite{stallkamp2011german} & 43 & 26,640& 12,630\\
    Country211~\cite{radford2021learning} & 211 & 42,200&21,100\\
    ImageNet~\cite{deng2009imagenet} & 1000 & 1.28M &50,000\\
    ImageNet-v2~\cite{recht2019imagenet} & 1000 & 1.28M&10,000\\
    \hline
  \end{tabular}}
  \label{tab:classification}
\end{table}

%% file: tables/multimodal_datasets.tex
\begin{table}[h]
  \centering
  \caption{The details of the datasets used in multi-modal tasks.}
  \resizebox{0.6\textwidth}{!}{
  \begin{tabular}{c|c|c}
    \hline
    Dataset & Task & Evaluation model \\
    \hline
    MSCOCO~\cite{lin2014microsoft} & image-text retrieval & BLIP\&CLIP\\
    Flickr30K~\cite{plummer2015flickr30k} & image-text retrieval & BLIP\&CLIP\\
    VQAv2~\cite{goyal2017making} & vision question answering & BLIP\\
    A-OKVQA~\cite{schwenk2022okvqa} & vision question answering & BLIP\\
    OK-VQA~\cite{marino2019ok} & vision question answering & BLIP\\
    NLVR~\cite{suhr2018corpus} & vision reasoning & BLIP\\
    MSCOCO~\cite{lin2014microsoft} & image captioning & BLIP\\
    NoCaps~\cite{agrawal2019nocaps} & image captioning & BLIP\\
    MSRVTT~\cite{xu2016msr} & video retrieval & BLIP\\
    \hline
  \end{tabular}}
  \label{tab:multimodal_datasets}
\end{table}

%% file: tables/parameter.tex
\begin{table}[h]
  \centering
  \caption{Hyperparameter configuration of MLLMs.}
  \begin{tabular}{c|c|c}
    \hline
    Model & Hyper-parameters & Value\\
    \hline
    \multirow{8}{*}{ Mini-GPT4-Vicuna13B} & max\_new\_tokens & 30\\
     & num\_beams & 1\\
     & do\_sample & True\\
     & min\_length & 1\\
     & top\_p & 0.3\\
     & repetition\_penalty & 1.0\\
     & length\_penalty & 1\\
     & temperature & 1.0\\
    \hline
    \multirow{3}{*}{Otter-Image-MPT7B} & max\_new\_tokens & 30\\
     & num\_beams & 1\\
     & no\_repeat\_ngram\_size & 3\\
    \hline
    \multirow{3}{*}{Qwen-VL-Chat} & max\_new\_tokens & 30\\
     & num\_beams & 1\\
     & do\_sample & False\\
     & min\_new\_tokens & 8\\
     & length\_penalty & 0\\
     & num\_return\_sequences & 1\\
    \hline
    \multirow{3}{*}{ LLaVA-v1.5-13B} & max\_new\_tokens & 30\\
     & num\_beams & 1\\
     & do\_sample & True\\
     & top\_p & None\\
     & temperature & 0.2\\
    \hline
  \end{tabular}
  \label{tab:parameter}
\end{table}